\pdfoutput=1
\documentclass{article}

\usepackage{microtype}
\usepackage{graphicx}
\usepackage{subfigure}
\usepackage{booktabs}
\usepackage{amsmath}
\usepackage{multirow}
\usepackage{amssymb}
\usepackage{diagbox}
\usepackage{tikz}

\usepackage{hyperref}
\usepackage[accepted]{icml2020}

\newcommand{\softmax}{\mathop{\mathrm{softmax}}}
\newcommand{\lstm}{\mathop{\mathrm{LSTM}}}
\newcommand{\bX}{\mathbf{X}}
\newcommand{\bx}{\mathbf{x}}
\newcommand{\bb}{\mathbf{b}}

\newcommand{\bH}{\mathbf{H}}
\newcommand{\bc}{\mathbf{c}}
\newcommand{\bQ}{\mathbf{Q}}
\newcommand{\bK}{\mathbf{K}}
\newcommand{\bV}{\mathbf{V}}
\newcommand{\bC}{\mathbf{C}}
\newcommand{\bk}{\mathbf{k}}
\newcommand{\bU}{\mathbf{U}}

\icmltitlerunning{MARVIN}

\begin{document}

\icmlsetsymbol{equal}{*}

\twocolumn[
\icmltitle{Multi-Agent Routing Value Iteration Network}

\begin{icmlauthorlist}
\icmlauthor{Quinlan Sykora}{equal}
\icmlauthor{Mengye Ren}{equal}
\icmlauthor{Raquel Urtasun}{}
\end{icmlauthorlist}

\icmlcorrespondingauthor{Quinlan Sykora}{quinlan.sykora@uber.com}
\icmlcorrespondingauthor{Mengye Ren}{mren3@uber.com}
\icmlcorrespondingauthor{Raquel Urtasun}{urtasun@uber.com}

\icmlkeywords{vehicle routing, planning, value iteration, multi-agent}
\vskip 0.3in
]

\printAffiliationsAndNotice{\icmlEqualContribution}

\vspace{-0.3in}
\begin{abstract}
In this paper we tackle the problem of routing multiple agents in a coordinated manner. This is a
complex problem that has a wide range of applications in fleet management to achieve a common goal,
such as mapping from a swarm of robots  and ride sharing. Traditional methods are typically not
designed for realistic environments which contain  sparsely connected graphs and unknown
traffic, and are often too slow in runtime to be practical. In contrast, we propose a graph neural
network based model that is able to perform multi-agent routing based on learned value iteration in
a sparsely connected graph with dynamically changing traffic conditions. Moreover, our learned
communication module enables the agents to coordinate online and adapt to changes more effectively.
We created a simulated environment to mimic realistic  mapping performed by autonomous vehicles with unknown minimum edge
coverage and traffic conditions;  our approach significantly outperforms traditional solvers
both in terms of total cost and runtime. We also show that our model trained with only two agents on
graphs with a maximum of 25 nodes can easily generalize to situations with more agents and/or nodes.
\footnote{Our code and data are released at \url{https://github.com/uber/MARVIN}}
\end{abstract}
\section{Introduction}
\label{intro}

As robots become ubiquitous, one of the fundamental problems is to be able to route a fleet or swarm
of robots to perform a task. Several approaches have been developed to try to solve this task.
The traveling salesman problem (TSP) is a classic NP-Hard~\cite{vrp} problem in computer science, wherein
an agent must visit a set of points while traveling the shortest possible distance.
The natural multi-agent generalization of this problem is known as the vehicle routing
problem (VRP)~\cite{vrp}, where multiple agents work in tandem to ensure that all points are visited exactly once.
Despite a plethora of classic
approaches to these problems~\cite{concorde,lkh3}, solvers are typically structured for offline
planning and generally are unable to adapt their solutions online. Furthermore, these solvers do not
incorporate online communication between the agents which is very desirable in many
practical settings. These are not necessarily weaknesses of the solutions but rather the
oversimplification of the problem definition itself.

Recently proposed deep learning methods have presented promising results in approximating solutions
with much faster runtime~\cite{pointer,combinatorialgraph,am,ean}. However, they are often tested on
simplistic planar graph benchmarks with limited exploration on multi-agent settings. Moreover, none
of these methods were designed towards dynamic environments where online communication can be very
beneficial.

\begin{figure}[t]
\begin{center}
\includegraphics[width=0.94\columnwidth,trim={3cm 0 6cm 10cm},clip]{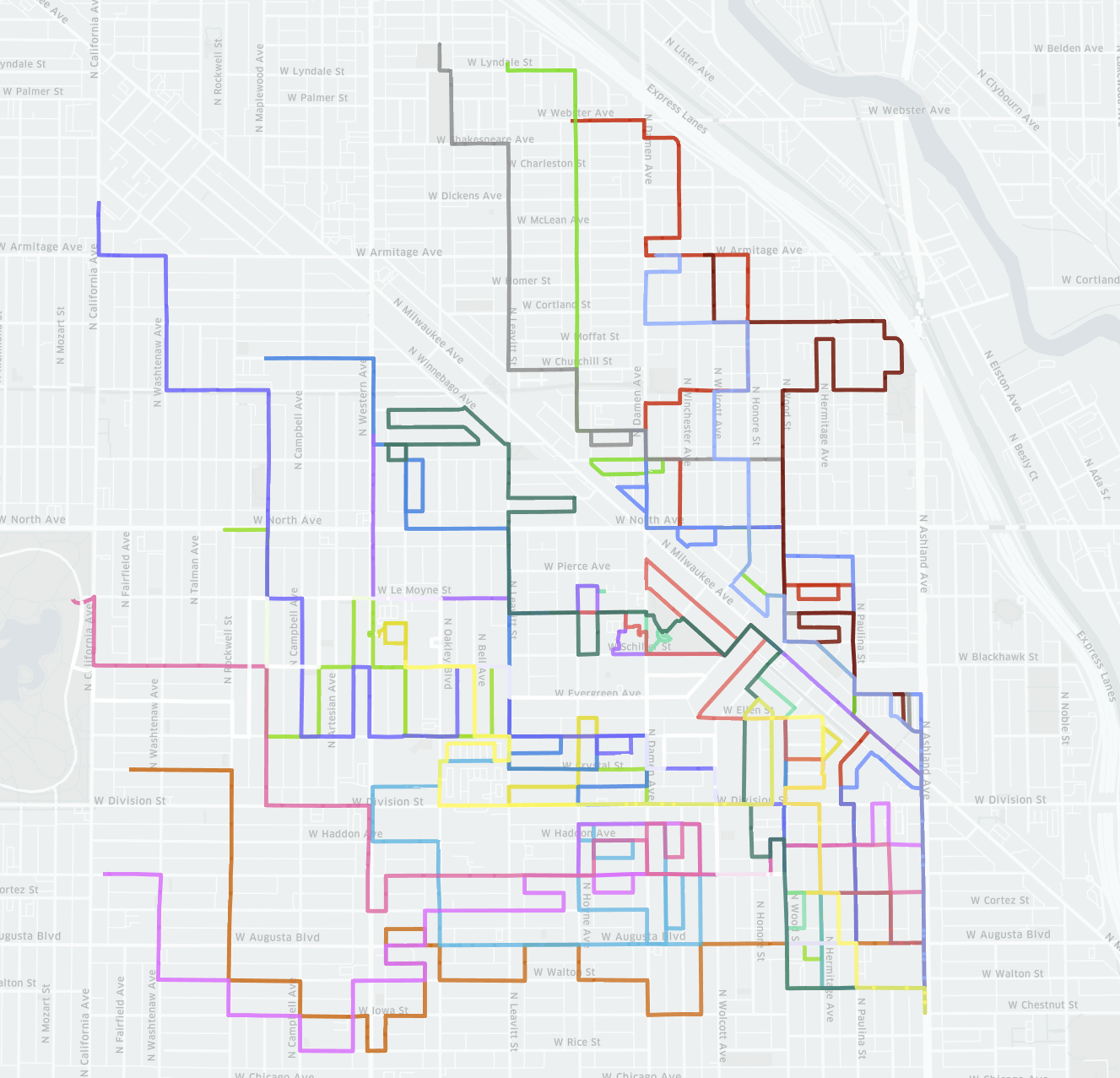}
\vspace{-0.15in}
\caption{A visualization of the route produced by a fleet of twenty vehicles using our proposed
algorithm. Colors denote different vehicle trajectories.}
\label{fig:teaser}
\end{center}
\vspace{-0.25in}
\end{figure}
We focus on the realistic {\it multi-agent autonomous mapping problem}: given a fleet
of vehicles, find the minimum total cost to map an urban region subject to traffic conditions,
such that each road in the city is traversed at least a certain number of times, where the number is
unknown a priori. This is a realistic setting for autonomous mapping as a region might need to be
recollected due to occlusions, heavy traffic, bad localization, sensor failure, bad weather
conditions, etc. Unfortunately neither VRP solutions nor existing deep learning methods perform well
in this difficult scenario.

In this paper we propose the Multi Agent Routing Value Iteration Network (MARVIN), a distributed
deep neural net tasked with the coordination of a swarm of vehicles to achieve a specific goal. In
particular, each agent performs local planning in a learned value iteration module which exploits
inter-agent communication through a novel learned communication protocol using the an attention mechanism.
As we focus on sparse road graphs, our second contribution is the use of a dense
adjacency matrix that encodes pairwise edge information to speed up information exchange and enable
more rich node encodings.

We demonstrate the effectiveness of our approach on real road maps extracted from 18 different
cities from around the world, using realistic traffic flow
simulation~\citep{macroscopicsim,continuum}.
To create our training and evaluation examples containing the
aforementioned realistic mapping challenges, we randomly subsample subgraphs on those cities, for
each node in each graph we then randomly sample the number of times it has to be covered. Note that this information will be unknown to the fleet, and will only discovered upon reaching this number.
We exploit total traversal time as our primary evaluation metric, and show that
our approach achieves better performance than state-of-the-art conventional VRP solvers~\citep{lkh3}, as well as recently proposed deep learning
models~\citep{am, ean, gvin}. Furthermore, MARVIN
generalizing well to the graph size and the number of agents guaranteeing a full
graph completion.

\section{Related Work}

In this section, we discuss previous attempts to solve vehicle routing problems. Background on
graph neural networks and value iteration networks is also
provided.

\paragraph{Vehicle routing problem:}

Existing VRP solvers can be broken down into two categories: conventional iterative solvers and deep
learning methods. Conventional solvers are usually iterative and designed to eventually converge to
the true optimal of the system~\citep{concorde, branchandcut,lkh3}.Some solvers are only designed
towards 2D planar graphs~\citep{spreadsheetvrp, gavrp, ilpvrp}. Structured for offline planning,
they are generally unable to adapt their solutions online. Moreover, they are not capable of any
online communication between agents to incorporate local observations.

In contrast to conventional solvers, deep learning methods have recently emerged as efficient
approximate solutions to combinatorial problems, thanks to the wide-spread success of attention
mechanisms~\citep{pointer,transformer} and graph neural networks~\citep{gcn,combinatorialgraph}.
Crucially, deep learning methods have powerful learning capabilities that can adapt easily to more
complex and realistic problem definitions. While some simply try to improve sub-problems of the VRP
task~\citep{marl,coopmasvrp, masterslave}, others produce end-to-end vehicle routes~\citep{am,
ean}. However, these deep learning solutions tend to assume that each node has a pair of 2D
coordinates that can be used to identify its global position, and edges are connected using
Euclidean distances, an unrealistic approximation of real road network graphs.
Furthermore, PointerNet~\citep{pointer,onlineroutenn} and Encode-Attend-Navigate(EAN)~\citep{ean},
two prominent deep learning TSP solvers, are restricted to the single agent domain, whereas
(AM)~\citep{am}, another deep learning solver which is able to operate in the multi-agent domain,
only does so by creating a route for one agent after another, and thus is unable to control the
exact number of agents being dispatched in each traversal.
Moreover, none of these methods were designed
to handle dynamic environments where one can benefit significantly from online communication.

\paragraph{Value iteration networks:}
Deep learning based methods have also shown promising performance in path planning. One classical
example is the value iteration network~\citep{vin}, which embeds structural biases inspired from
value iteration~\citep{bellman} in a neural network. Gated path planning networks~\citep{gppn}
changed the max-pooling layer with a generic long short-term memory (LSTM)~\citep{lstm}
significantly improving training stability which helps extend the number of iterations. These
networks can naturally be translated to a graph domain by replacing the transitions with the edges
in the graph, as is shown in the generalized value iteration network (GVIN)~\citep{gvin}. However,
they are developed to solve simple path planning environments such as 2D mazes and small graphs with
weighted edges, and Dijkstra's shortest path algorithm is already efficient and effective at solving these
 problems.
Compared to the design of GVIN, our method features a dense adjacency matrix that is very
effective at solving sparse graph coverage problems, where long range information exchange is
needed.

\paragraph{Graph neural networks:}
Graph neural networks~\cite{gnn,gnnsurvey} provide a way to learn graph representations that are
both agnostic to the number of nodes in the graph and permutation invariant in the local
neighborhood. Information from node neighborhood can be aggregated using graph
convolutions~\citep{gcn}, recurrent neural networks~\cite{ggnn}, and more recently via attention
mechanisms~\cite{gtn,gat}. Graph attention modules also appear in deep learning based VRP/TSP solvers such
as AM~\cite{am} and EAN~\cite{ean}.
Inspired by prior literature, we make use of graph attention in two ways: 1) a
map-level road network augmented with graph attention within the planning module of each agent, and 2) an agent-level attention
to aggregate messages received from other agents.

\paragraph{Multi-agent communication:}
Traditional multi-agent communication in robotics has focused on heuristic and algorithmic
approaches to improve communication efficiency~\citep{dynamicroute, maretrieval, commefficiency}. In
contrast, CommNet~\citep{commnet} and its natural extensions ~\citep{learningCoordination} demonstrated that a swarm of agents can autonomously
learn their own communication protocol. This has led to a focus on the nature of learned language
protocols. Some studies \citep{commnet, coop, attcomm} propose ways to combine information
among agents. \citet{commnet} use a simple summation across the messages, whereas \citet{attcomm}
leverage the attention mechanism to identify useful information. Other works focus more on the
difference between cooperative swarms, greedy individuals, and competing swarms with learned
communication~\citep{emergence, multiagentrl}. Finally, there is a large body of work on the
scalability of robotic swarms~\citep{graphpolicygrad}, and the necessity of explicit communication
to infer the actions of other agents~\citep{macontrol}.

\begin{figure*}[t]
\begin{center}
\includegraphics[width=0.975\textwidth,trim={0 7.5cm 2cm 0.5cm},clip]{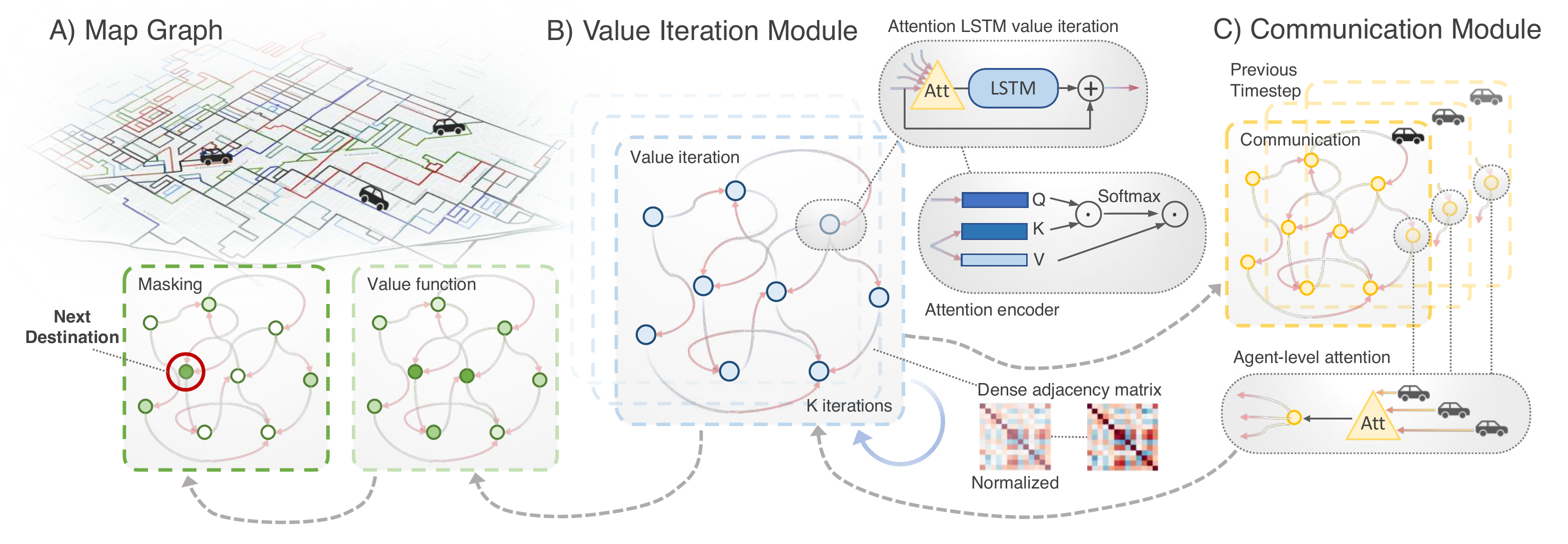}
\vspace{-0.15in}
\caption{\textbf{Our proposed multi-agent routing value iteration network}:
\textbf{A)} The map is represented as a graph and each node has some local observation features;
\textbf{B)} Each agent operates its own value iteration network. It uses an attention-based LSTM on
each graph node to exchange information. The LSTM runs for $k$ iterations and can be decoded into a
value function for selecting next destination. \textbf{C)} Inter-agent communication is implemented
with an attention mechanism over all the incoming messages, and the output is fed as additional
channels of inputs to the value iteration module.}
\label{fig:mainfig}
\end{center}
\vspace{-0.2in}
\end{figure*}
\section{Problem Definition}

In this section we first provide a precise definition of the multi-agent mapping problem. We then
propose in the next section a decentralized deep neural network for coordinating a fleet of vehicles
to solve this mapping problem. Formally, given a strongly connected directed graph $G(V,E)$
representing the road connectivity, we would like to produce a routing path for a set of $L$ agents
$\{p^{(i)}\}_{i=1}^L$ such that each vertex $v$ in $V$ is covered $M_v$ times in total across all
agents. We consider the real-world setting where 1) $M_v$ is unknown to all agents until the
number has been reached (i.e., only success/failure is revealed upon each action) and 2) only local
traffic information can be observed.

We consider a decentralized setting, where each agent gathers local observations and
information communicated from other agents, and outputs the route it needs to take in the next step.
Here we assume that each agent can broadcast to the rest of the fleet as this is possible with
today's communication technology. We also constrain the policy of
each agent to be the same, making the system more robust to failure.

Let $a_t^{(i)}$ be the routing action taken by agent $i$ at time $t$, indicating the next node to
traverse. We define a {\it route} as the sequence of actions $p^{(i)} = [a_0^{(i)}, \dots,
a_N]$, where each action represents an intermediate destination.
We refer the reader to
Table~\ref{tab:notation} for our notation.

The policy of a single agent $i$ can be formulated as a function of 1) the road network graph $G$; 2) local
environment observation $o_t^{(i)}$; 3) the communication messages sent by other agents
$\{\bc_t^{(j)}\}$; and 4) the state of the agent $s_t^{(i)}$.
Thus,
\begin{align}
\{a_t^{(i)}, \bc_t^{(i)}\} = f(G, o_t^{(i)}, \{\bc_{t-1}^{(j)}\}_{j=1}^L; s_t^{(i)}),
\end{align}

\begin{table}[t]
\centering
\begin{small}
      \resizebox{0.8\columnwidth}{!}{%
\begin{tabular}{ccc}
\toprule
Symbol                 & Description                                              \\
\midrule
$t$                    & Current timestep                                         \\
$G$                    & Map graph                                                \\
$L$                    & Number of agents                                         \\
$n$                    & Number of graph nodes                                    \\
$f$                    & Routing + communication policy                           \\
$\pi$                  & Routing policy                                           \\
$F$                    & Time cost given a route $p$                              \\
$o_t^{(i)}$            & Observation by agent $i$ at time $t$                     \\
$s_t^{(i)}$            & State of agent $i$ at time $t$                           \\
$a_t^{(i)}$            & Action taken by agent $i$ at time $t$                    \\
$\bc_t^{(i)}$          & Message vector sent by agent $i$ at time $t$             \\
$M_v$                  & Number of times node $v$ needs to be covered             \\
$\bX^{(i,k)}_t$        & Agent $i$'s node features at $k$-th value iteration      \\
$\bU_i$                & The input communication features for agent i             \\
\bottomrule
\end{tabular}}
\end{small}
\vspace{-0.1in}
\caption{Notation}
\label{tab:notation}
\vspace{-0.2in}
\end{table}

Assuming that a traffic model $F$ produces the time needed to traverse a route, we would like our
multi-agent system to minimize the following objective:
\begin{align}
\min_{p^(i)}      &&& \sum_{i=1 \dots L} F(p^{(i)}), \\
\text{subject to} &&& \nonumber \sum_i M(p^{(i)}, v) \ge M_v, \ \ \forall v,
\end{align}
where $M(p, v)$ is the number of times node $v$ is visited in a route $p$.

\section{Multi-Agent Routing Value Iteration Network}

In this section, we describe our proposed approach to the multi-vehicle routing problem. Note that the model is running locally in each individual agent,
as this makes it scale well with the number of agents and be more robust to failures. There are two
main components of our approach. First, the \textbf{communication module} (Fig.~\ref{fig:mainfig}C) works asynchronously to save
messages sent from other agents in a temporary memory unit, and retrieves the content based on an
attention mechanism at the agent level.
Each time an agent needs to select a new destination,
this information is then sent to the value iteration module for
future planning. Second, the \textbf{value iteration module} (Fig.~\ref{fig:mainfig}B)
runs locally on each agent
and iteratively estimates the value of traveling to each node in the
road network graph for its next route
(Figure~\ref{fig:mainfig}A).
Then an attention LSTM planning module iteratively refines the node
features for a fixed number of iterations, and outputs the value function for each node. The node
with the highest value will be considered as the next destination for the agent. We now describe the
value iteration module followed by the communication module.

\begin{table}[t]
\centering
\begin{small}
\resizebox{0.65\columnwidth}{!}{%
\begin{tabular}{ccc}
\toprule
Name                                   & Dim. & Type             \\
\midrule
Sum of in/out edge weights             & 2    & float            \\
\# of in/out edges                     & 2    & integer          \\
Agent at $v$                           & 1    & binary           \\
$v$ is unexplored                      & 1    & binary           \\
$v$ is fully covered                   & 1    & binary           \\
Dist. from cur. pos. to $v$            & 1    & float            \\
Traffic at $v$                         & 1    & float            \\
$v$ is adjacent                        & 1    & binary           \\
Communication vector                   & 16   & float            \\
\bottomrule
\end{tabular}
}
\end{small}
\caption{Graph input feature representation for node $v$}
\vspace{-0.25in}
\label{tab:node_feature}
\end{table}

\subsection{Value Iteration Module}
Our model operates on a strongly connected graph $G(V,E)$ representing the topology of the road
network. As shown in Fig.~\ref{fig:mainfig}, each street segment forms a node in the graph, and the goal for
each agent is to pick a node to be its next destination.
Given some
initial node features, our approach refines them for a fixed
number of iterations of the graph neural network, decodes the features into a scalar value function for
each node, and then selects the node with the maximum value to be our next destination (see
Fig.~\ref{fig:mainfig}). We now provide more details on each of these steps.

Let $\bX = \{\bx_1, \bx_2, ... , \bx_n\}$ be the set of initial node feature vectors with $n$ being the
total number of nodes and let $\bU = \{u_1, u_2, ... , u_n\}$ represent the input communication node features.
We encode the node input features (see Table~\ref{tab:node_feature}) through a linear layer to serve as
initial features for the value iteration network:
\begin{align}
\bX^{(0)} &= (\bX \mathbin\Vert \bU) W_{\mathrm{enc}} + \bb_\mathrm{enc}.
\end{align}

At each planning iteration $t$, we perform the following iterative update through an LSTM with an
attention module across neighboring nodes:
\begin{equation}
  \bX^{(k+1)} = \bX^{(k)} + \lstm(\mathrm{Att}(\bX^{(k)}, A); \bH^{(k)}),
\end{equation}
for $t=1 \dots K$ and $K$ is the total number of value iteration steps. $\bH^{(t)}$ is the hidden
state of the LSTM, which contains one state vector per node, and $A$ is the adjacency matrix.
As opposed to conventional methods where the binary adjacency matrix is used as the primary input to
the network, we use the Floyd-Warshall algorithm to compute the dense
distance matrix as an explicit input in our architecture, thereby ensuring
more meaningful information can be utilized by our model. In particular, the matrix produced by the Floyd-Warshall
algorithm encodes the pairwise minimum path
distance between any pair of nodes, $D_{i,j} = d(v_i,v_j)$, which we normalize to form our dense
adjacency matrix.
$A = \frac{D - \mu}{\sigma}$, where $\mu$ is the element-wise mean of $D$, and
$\sigma$ is the element-wise standard deviation. As shown in our experiments, using our dense adjacency
matrix results in significantly better planning than the binary connectivity matrix of GVIN~\cite{gvin}).

\paragraph{Graph attention layer:} Information exchange on the graph level happens in the attention
module ``$\mathrm{Att}$'' which is a transformer layer~\cite{gtn}, that takes in the node features
and the adjacency matrix, and outputs the transformed features. Specifically, we first compute the
key, query, and value vectors for each node:
\begin{align}
  \bQ^{(k)} &= \bX^{(k)} W_q + \bb_q, \\
  \bK^{(k)} &= \bX^{(k)} W_k + \bb_k, \\
  \bV^{(k)} &= \bX^{(k)} W_v + \bb_v.
\end{align}
We then compute the attention between each node and every other node to create an attention matrix
$A_{\text{att}} \in \mathbb{R}^{n \times n}$,
\begin{equation}
  A_{\text{att}} = \bQ^{(k)} \bK^{(k)\top}.
\end{equation}
We combine the graph adjacency matrix $A$ with the attention matrix $A_{\text{att}}$ to represent
edge features as follows:
\begin{equation}
  \tilde{A}^{(k)} = \softmax(g(A_{\text{att}}^{(k)}, A)),
\end{equation}
where $g$ is a learned multi-layer neural network.

The new node values are computed by combining the values produced by all other nodes
according to the attention in the fused attention matrix. The output of the graph attention layer is
then fed to an LSTM module:
\begin{equation}
  \bX^{(k+1)} = \bX^{(k)} + \lstm(\tilde{A}^{(k)} \mathbf{V}^{(k)}; \bH^{(k)}).
\end{equation}

This full process is repeated for a fix number of iterations $k=1, \cdots, K$ before decoding. 

\paragraph{Value masking and decoding:}
After iterating the attention LSTM module for $K$ iterations, we use a linear layer to project the
features into a scalar value function for each node on the graph. We mask out the value of all nodes
that no longer need to be visited since they have been fully mapped, and take a softmax over
all remaining nodes to get the action probabilities
\begin{align}
\pi(a_i; s_i) = \softmax(\bX^{(K)} W_{\mathrm{dec}} + \bb_{\mathrm{dec}}).
\end{align}
Finally, we take the node that has the maximum probability value to be the next destination. The full route
will be formed by connecting the current node and the destination by using a shortest path algorithm on the weighted
graph. Note that the weights are intended to represent the expected time required to travel from one road segment to the next,
and therefore are computed by dividing the length of the street segment by the average speed of the vehicles traversing it.

\subsection{Communication Module}
Due to the partial observation nature of our realistic problem setup (\textit{e.g.,} traffic and
multiple revisits), it is beneficial to let the agents communicate their intended trajectories, thereby
encouraging more collaborative behaviours.
Towards this goal, our proposed model also features an
attention-based communication module, where now attention is performed over the agents, not the street segments.
Whenever an agent performs an action, it uses
$\bX^{(K)}$, the final encodings of the value iteration module,
to output the communication vector
: $\bc^{(i)}$, which is then broadcasted to all agents. We express the communication vector as a set of node
features in order to reflect the structure of the street graph environment.
he most recent communication vector
from each sender is temporarily saved on the receiver end. When an agent decides to take a new
action, it applies an agent-level attention layer to aggregate information from its receiver inbox.

Let $\bC_{\mathrm{in}} = \{\bc^{(1)}, \dots, \bc^{(L)}\} \in \mathbb{R}^{L \times nd}$,
be the messages that an agent receives from other agents concatenated together, where $L$ is the number of agents,
$n$ is the number of nodes and $d$ is the features dimension. The agent transforms the communication vectors
to produce a query and a value vector:
\begin{align}
  \bQ_{\mathrm{comm}} &= \bC_{\mathrm{in}} W_{q,{\mathrm{comm}}} + \bb_{q,\mathrm{comm}}, \\
  \bV_{\mathrm{comm}} &= \bC_{\mathrm{in}} W_{v,{\mathrm{comm}}} + \bb_{v,\mathrm{comm}}.
\end{align}
The communication vector last outputted by this given agent is also called upon to produce a key
vector:
\begin{equation}
  \bk_{i,{\mathrm{comm}}} = \bC_{\mathrm{in}, i} W_{k,\mathrm{comm}} + \bb_{k,\mathrm{comm}}.
\end{equation}
This key vector is then similarly dotted with the query vectors from all other agents to form a
learned linear combination of the communication vectors from all the other agents.
We can then compute the aggregated communication as
$\bU_{i}$:
\begin{align}
\bU_{i} &= \sum_j \alpha_{i,j} \mathbf{V}_j, \\
\mathbf{\alpha}_i &= \softmax{(\mathbf{Q_{\mathrm{comm}}} \mathbf{k}_{i,{\mathrm{comm}}})}.
\end{align}
$\bU_{i}$ will then be used as part of the node feature inputs to the value iteration module for the
next step.

\subsection{Learning}
Our proposed network can be trained end-to-end using either imitation learning or reinforcement
learning. Here we explore both possibilities. For imitation learning, we assume there is an
oracle that can solve these planning problems. Note that this relies on a fully observable
environment, and oftentimes the oracle solver will slow down the training process since we generate
a training graph for each rollout.
Alternatively, we also consider training the network using
reinforcement learning, which is more difficult to train but directly optimizes the final objective.
We now describe the learning algorithms in more details.

\paragraph{Imitation learning (IL):}
To generate the ground-truth $a^\star$ that we seek to imitate, we firstly provide an LKH3 solver
with global information about each problem to solve as a fully observed environment. Based on the
ground-truth past trajectory, each agent tries to predict the next move $a$. We train the agent
using ``teacher-forcing'' by minimizing the cross entropy loss for each action, summing across the
rollout. In teacher forcing, the agents are forced to perform the same actions as the ground truth rollout at
each timestep, and are penalized when their actions do not match that of their ``teacher''. The loss is averaged across a mini-batch.
\begin{align}
L = - \mathbb{E}[\sum_{t,i} \log \pi(a_t^{(i)\star}; s_t^{(i)}) ],
\end{align}
where $\pi(a; s)$ denotes the probability of taking action $a$ given state $s$.

\paragraph{Reinforcement learning (RL):}
While imitation learning is effective, expert demonstration may not always be available for realistic environments. Instead, we can use
reinforcement learning. We use REINFORCE~\cite{reinforce} to train
the network using episodic reinforcement learning, and set the negative total cost of the fully
rolled out traversal to be the reward function, normalized across a mini-batch.
\begin{align}
r &= - \sum_{i} F(p^{(i)}), \quad
\tilde{r} = (r - \mu_r) / \sigma_r, \\
L &= - \mathbb{E}_{\pi}\tilde{r}, \ \
\nabla L = -\mathbb{E}_{\pi}[
\tilde{r} \sum_{t, i} \nabla \log \pi (a_t^{(i)}; s_t^{(i)} )
].
\end{align}

\section{Autonomous Mapping Benchmark}
In this section we describe our novel autonomous mapping benchmark. The dataset contains
22,814 directed road graphs collected from 18 cities around the world from different continents.
We refer the reader to Table~\ref{tab:stats} for statistics of our dataset. We use a separate city for testing
purposes and 10\% of the training set for validation. We also augment this benchmark with realistic
traffic conditions and realistic mapping challenges. These extra challenges fall into the following
categories: random revisits, realistic traffic and asynchronous execution.

\vspace{-0.1in}
\paragraph{Random revisits:}
When mapping a road in the real world, it is possible that our initial mapping attempt
could fail, due to occlusion, sensor uncertainty, etc. Therefore we would have to revisit that
street an unknown number of times before it is fully mapped. In order to simulate this,
at the beginning of each run, we assign each node in the street graph a hidden variable
that corresponds to how many times it will have to be visited before it is fully mapped.
During training, we sample this value uniformly from one to three. We also randomly sample
this value uniformly from one to three during evaluation, except for when we specifically
test for an alternate distribution (see Table~\ref{tab:multipass}).

\vspace{-0.1in}
\paragraph{Traffic simulation:}
We also simulate unknown traffic congestion for each street. To find the equilibrium
congestion at each node, we use the flow equations proposed in~\citet{macroscopicsim}. This
method simulates traffic as a flow problem wherein we wish to maximize the total movement of
vehicles given a set of junction constraints. We use the number of incoming and outgoing lanes
multiplied by the speed limit of those lanes to establish the flow constraints, and initialize
the congestion randomly using a uniform distribution from 0 to 1. Once we find an approximation
for the equilibrium congestion, following ~\citet{continuum} we define the velocity at each
street $v$ to be:
$v = v_{\max} * (1 - \rho ^ {\gamma})$,
where $v_{\max}$ represents the speed limit of that road, $\rho$ is the traffic congestion on that
road and $\gamma$ is a hyperparameter (that we set to 3) that helps smooth out the effect of traffic.
The effect of this is that whenever an agent travels to a particular node, the cost of performing
this traversal is increased by $\frac{1}{(1 - \rho ^ {\gamma})}$.
We cap this factor to a maximum
value of 4 to ensure that the cost of traveling to nodes with maximum congestion does not extend to
infinity. This allows the cost of an edge traversal to vary between 1 to 4 times its
original value depending on the equilibrium congestion. Note that the congestion value of each node
is unknown until the node is visited by an agent.

\begin{table}[t]
\begin{small}
\begin{center}
\begin{tabular}{cccc}
\toprule
Set     & \# Graphs & \# Nodes    \\
\midrule
Train  & 22,814    & 420,452  \\
Test   & 373       & 14,284      \\
\bottomrule
\end{tabular}
\end{center}
\end{small}
\vspace{-0.1in}
\caption{Realistic autonomous mapping benchmark statistics}
\label{tab:stats}
\vspace{-0.2in}
\end{table}
\paragraph{Asynchronous execution:}

During the training phase the agents act in a synchronous manner, where
each agent is
called sequentially to perform an action until the graph has been entirely mapped.
This however does not take into account the time required to perform each action.
 During the
evaluation phase, we instead simulate the time required to complete each action. Agents
therefore act in an asynchronous way based on how long each action takes to complete.

\section{Experiments}
\subsection{Implementation Details and Baselines}
Each graph node has 16 communication channels for each agent. Similarly, the dimension of the
encoding vectors is 16. For combining the dot-product attention with the distance matrix, a 3-layer
[16-16-16] MLP with ReLU activation is used. When training, we set the learning rate of our model to
be 1e-3 using the Adam optimizer, with a decay rate of 0.1 every 2000 epochs. We train our model for
5000 epochs. We use a batch size of 50 graphs, each of which has up to 25 nodes. We train our
network with two agents only and evaluate with settings varying from one to nine agents.

\begin{figure*}[t]
\begin{center}
\centerline{\includegraphics[width=0.95\textwidth,trim={1cm 1cm 1cm 0}]{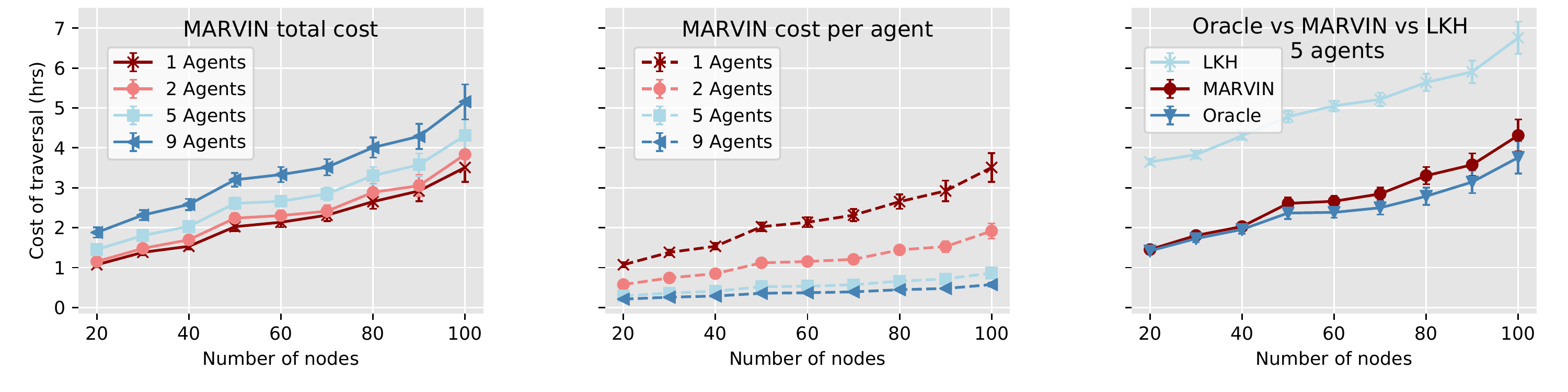}}
\caption{Number of agents performing traversal and corresponding cost of the traversal
(trained using RL)}
\label{fig:scalability}
\end{center}
\vspace{-0.25in}
\end{figure*}
  \begin{table*}[t]
  \begin{center}
    \centering
      \resizebox{0.9\textwidth}{!}{%
      \begin{tabular}{c|ccc|ccc|ccc|ccc}
      \toprule
                  & \multicolumn{3}{c}{n = 25, 1 Agent}                & \multicolumn{3}{c}{n = 25, 2 Agents}                   & \multicolumn{3}{c}{n = 50, 2 Agents}            & \multicolumn{3}{c}{n = 100, 5 Agents}              \\
      Method      & Cost          & Gap             & Runtime          & Cost           & Gap             & Runtime             & Cost          & Gap             & Runtime       & Cost          & Gap             & Runtime          \\
      \midrule
      \midrule
      Oracle      & 1.16          & 0.00\%          & 71.3             & 1.28           & 0.00\%          & 438                 & 1.85          & 0.00\%          & 902           & 3.19          & 0.00\%          & 2430             \\
      \midrule
      Random      & 4.45          & 284.9\%         & \textbf{3.15}    & 4.47           & 249.4\%         & \textbf{1.50}       & 8.25          & 345.2\%         & \textbf{1.82} & 18.9          & 492.2\%         & \textbf{2.83}    \\
      Greedy      & 2.12          & 73.7\%          & 3.37             & 2.33           & 81.8\%          & 2.11                & 3.55          & 91.5\%          & 3.57          & 10.4          & 227.0\%         & 22.3             \\
      LKH3        & 1.26          & 8.84\%          & 71.2             & 1.80           & 40.5\%          & 438                 & 2.54          & 37.3\%          & 902           & 6.14          & 92.5\%          & 2430             \\
      \midrule
      GVIN        & 1.37          & 18.8\%          & 52.5             & 1.48           & 15.9\%          & 44.2                & 2.45          & 32.1\%          & 63.4          & 5.41          & 69.6\%          & 48.6             \\
      GAT         & 1.53          & 32.5\%          & 43.0             & 1.56           & 21.6\%          & 29.1                & 2.58          & 39.7\%          & 38.0          & 5.43          & 70.2\%          & 38.2             \\
      AM          & 4.90          & 322.4\%         & 161              & -              & -               & -                   & -             & -               & -             & -             & -               & -                \\
      EAN         & 2.89          & 145.8\%         & 212              & -              & -               & -                   & -             & -               & -             & -             & -               & -                \\
      MARVIN (IL) & 1.37          & 18.0\%          & 62.8             & 1.42           & 11.3\%          & 66.6                & 2.21          & 19.0\%          & 71.5          & \textbf{4.36} & \textbf{36.7\%} & 72.8 \\
      MARVIN (RL) & \textbf{1.25} & \textbf{8.17\%} & 62.8             & \textbf{1.32}  & \textbf{2.87\%} & 56.6                & \textbf{2.12} & \textbf{14.5\%} & 71.4          & 4.62          & 44.9\%          & 72.8 \\
      \bottomrule
      \end{tabular}
      }
      \caption{Average graph traversal cost on realistic graphs; Time cost in hours; Runtime in milliseconds.}
        \label{tab:1}
        \vspace{-0.2in}
        \end{center}
    \end{table*}

We compare our approach with the following baselines.
\paragraph{Random:} This baseline consists of visiting each node that has not been completely mapped
yet in a random order.

\vspace{-0.1in}
\paragraph{Greedy:} Each agent visits the closest node that still needs to be mapped. It assumes
that the agents are able to communicate which nodes have been fully mapped.

\vspace{-0.1in}
\paragraph{LKH3:} represents the best performance of the iterative solver given limited information.
We first allow the solver ~\citep{lkh3} to calculate the optimal path for covering each
 node exactly once. Then, the solver calculates a new optimal path over all the remaining
nodes that must be mapped. This is repeated until all nodes have been fully mapped. In essence, the
solver performs VRP traversals until all nodes have been visited the desired number of times.

\vspace{-0.1in}
\paragraph{GVIN:} The Generalized Value Iteration Network~\citep{gvin} uses a GNN to propagate
values on a graph. While the original implementation does not integrate communication between
multiple agents, we enhanced GVIN with our attention communication module for fair comparison. This
model was trained with imitation learning to achieve best performance.

\vspace{-0.1in}
\paragraph{GAT:} Graph Attention Networks~\citep{gat} are similar to our method, in that they
exchange information according to attention between two nodes in order to convey complex
information. However, standard GAT architectures do not encode the distance matrix information and
instead assume all edges have an equal weight, limiting their capabilities.
While GATs are not necessarily designed to solve the TSP or VRP problems, they remain
one of the state-of-the-art solutions for graph and network encoding.

\vspace{-0.1in}
\paragraph{AM:} The Attention Model~\citep{am} has been used as a deep learning VRP solver. Since
it has no natural way to encode the information in the adjacency matrix, we add a Graph Convolution Network (GCN) 
encoding module at the beginning to perform this encoding and use imitation learning for training. The
modification that the authors suggest for the AM algorithm to allow it to perform a VRP traversal
does not allow for dynamic route adaptation, as is required in our environment. We therefore only
evaluate this method in the single agent scenario.

\vspace{-0.1in}
\paragraph{EAN:} Encode-Attend-Navigate~\citep{ean} is a deep learning TSP solver designed very
similarly to AM, but with a slightly different architecture. We enhance it with an additional GCN
module at its beginning to encode the adjacency matrix in the same fashion as with AM. This model
was trained with imitation learning as well.

\vspace{-0.1in}
\paragraph{Oracle:} This is the upper bound performance that an agent could have possibly achieved
if given global information about all the hidden states. This solution is found by providing the
LKH3 solver with details about all the hidden variables, and solving for the optimal plan. We transform
the adjacency matrix by increasing the edge weights of the nodes effected by congestion, and by
duplicating the nodes that will require multiple passes to be fully mapped.

\begin{figure*}[t]
\centering
\begin{small}
\vspace{-0.1in}
\begin{tabular}{llll}
(a) & (b) & (c) & (d)\\
\includegraphics[height=4.1cm,trim={0.2cm 0 0.4cm 0},clip]{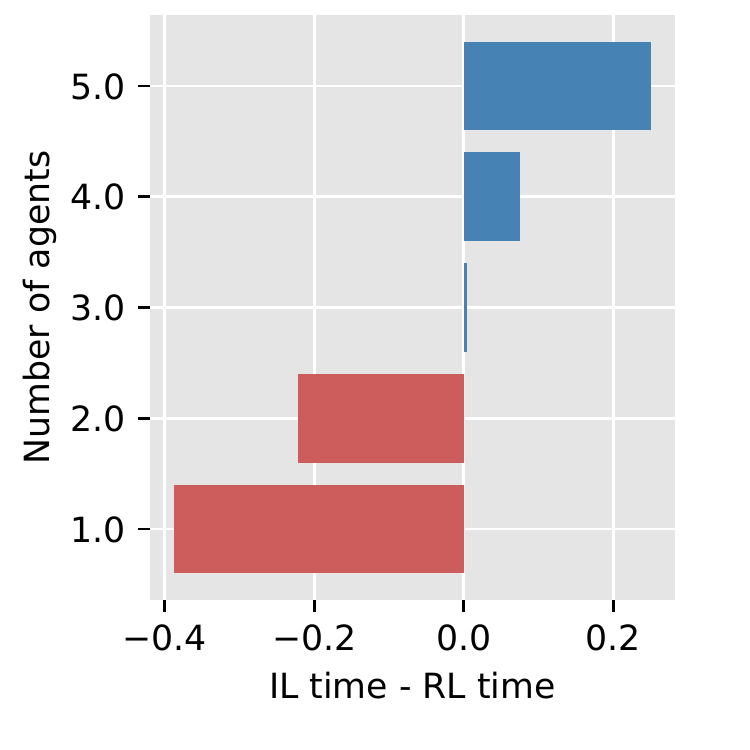} &
\includegraphics[height=4.1cm,trim={0.2cm 0 0.8cm 0},clip]{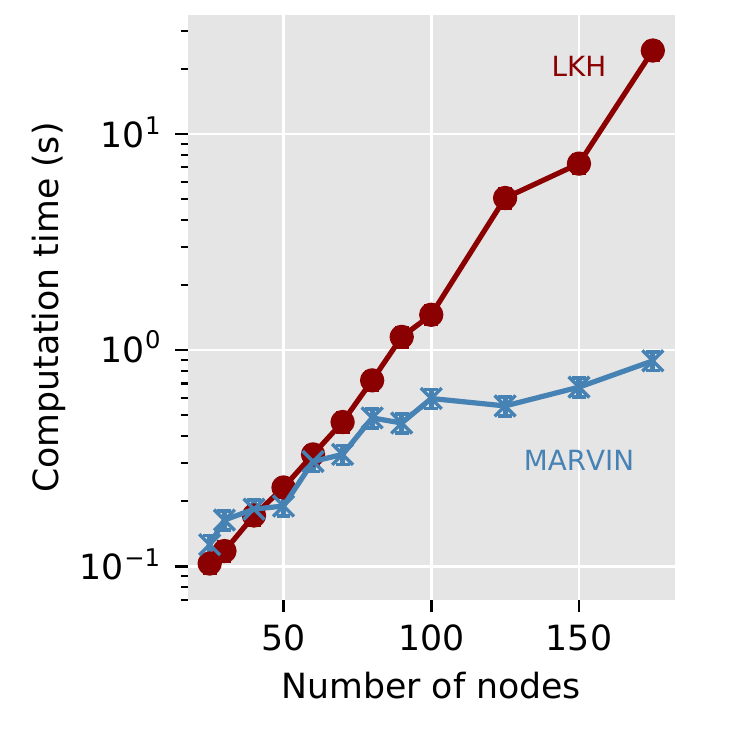} &
\includegraphics[height=4.1cm,trim={0.2cm 0 0.8cm 0},clip]{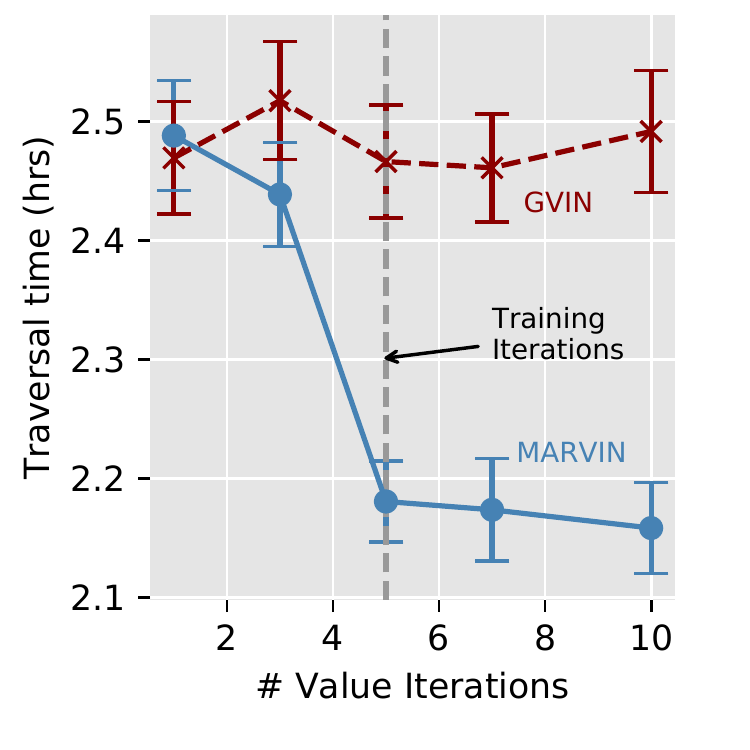} &
\includegraphics[height=4.1cm,trim={0.0cm 0 0cm 0},clip]{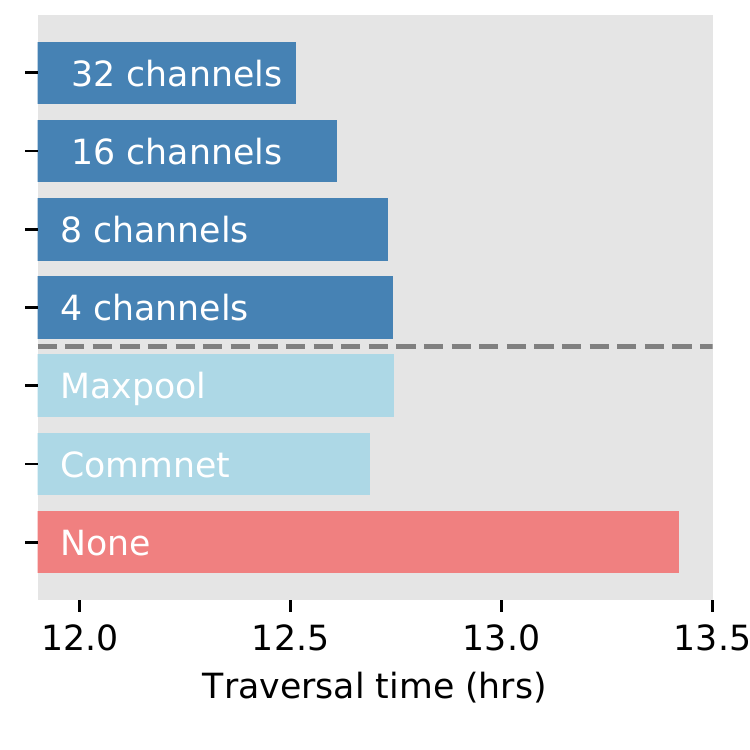}\\
\end{tabular}
\end{small}
\vspace{-0.1in}
\caption{\textbf{(a) IL vs. RL:} Comparison of imitation learning and reinforcement learning on different number of agents;
\textbf{(b) Runtime:} Comparison MARVIN's runtime to that of the LKH solver;
\textbf{(c) No. iterations in the VIN module:} Evaluation of how the number of value iterations has on performance, and how the number of
    iterations generally scales for other value iteration models (GVIN);
\textbf{(d) Communication module design:} Comparison of our communication protocol to other communication protocol alternatives.
}
\label{fig:all}
\end{figure*}
\subsection{Results}
\vspace{-0.1in}
\paragraph{Comparisons to Baselines:} As shown in Table~\ref{tab:1}, our method has the best
performance across different numbers of agents and graph sizes. Notably, under 25 nodes and two
agents, which is the training setting, our method with RL achieves a total cost that is within 3\%
from that of the oracle. We found our model trained with both reinforcement learning and imitation
learning outperforms all competitor models. Overall, the model shows impressive generalization to
more agents and larger graph size, since we limit the training of the model with two agents and 25
nodes. We also note that deep learning-based solvers that perform well in the more traditional TSP
domain (AM, EAN) are unable to generalize well to our realistic benchmark. Specifically, the low
performance of these deep learning solvers can be attributed to their inability to cope with mapping
failures and nodes requiring multiple passes, which in turn is the result of their architecture
not being structured for this problem formulation.

\vspace{-0.1in}
\paragraph{Load distribution:} The cost of performing each traversal is very evenly spread out
among all of the agents. The maximum gini coefficient for two agents observed on our evaluation set
was 0.169, with the mean coefficient being around 0.075.

\vspace{-0.1in}
\paragraph{Scale to number of agents and graph size:} One
of the primary focuses of our work is to develop a model that scales well with the number of agents
and the size of the graph. Therefore, we evaluated our model's performance on increasingly large
graphs and observed how the performance varied with the number of agents. As shown in Fig.~\ref{fig:scalability},
 the total cost increases marginally when we increase the total number of agents,
indicating good scalability in this respect, and that our method performs much better than the
current state of the art, as is represented by LKH.

We notice that models trained with RL appear to be able to generalize better when dealing with a
larger number of agents, shown in Fig.~\ref{fig:all}A. This could be explained by the fact that
supervised learning tries to exactly mimic the optimal strategy, which may not carry over when
dealing with more agents, and therefore generalizes less effectively.

\begin{figure}[t]
\begin{center}
\includegraphics[width=0.94\columnwidth]{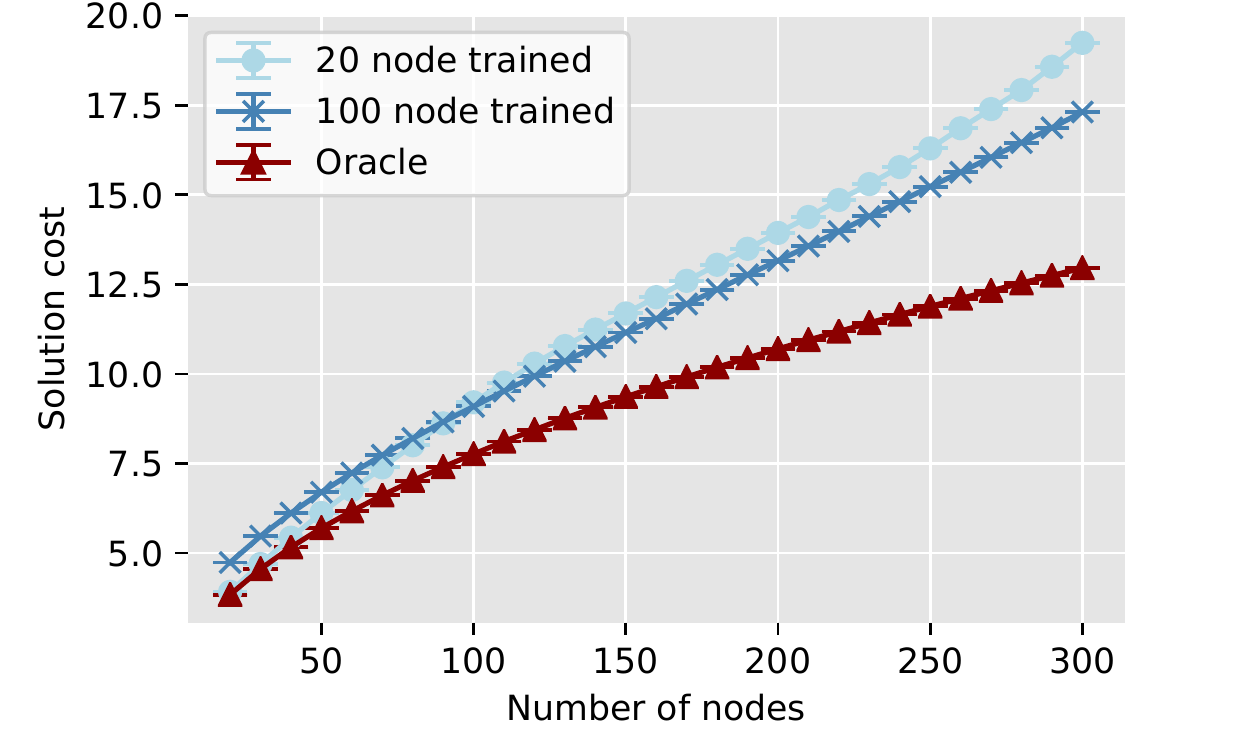}
\vspace{-0.2in}
\caption{The average traversal time (hrs) relative to the number of nodes in the graph
for policies trained exclusively on 25 node graphs and 100 node graphs
}
\vspace{-0.25in}
\label{fig:retrained}
\end{center}
\end{figure}

We also evaluate how a model trained on only toy graphs with 25 nodes compares with a graph trained
only on toy graphs with 100 nodes.
As shown in Fig.~\ref{fig:retrained} ,the 100 node model scales much
better on larger graphs, but that 25 node version of the model is able to perform much closer to
true optimal when acting on smaller graphs.


\vspace{-0.1in}
\paragraph{Runtime:} An advantage that deep learning solutions have over conventional
solvers is their runtime. We compare the average runtime of our model versus that of the LKH3
solver in Fig.~\ref{fig:all}B. Note that our model is significantly faster than LKH3 on large
scale graphs with over 100 nodes.

\vspace{-0.1in}
\paragraph{Robustness to distribution shift:} We also evaluate the generalizability of our model to
different ``multiple pass'' distributions. In order to simulate realistic mapping failures, each
node must be revisited an unknown number of times before we say that it has been completely mapped.
The ``multiple pass'' distributions is the distribution from which these numbers are selected.
Shown in Table~\ref{tab:multipass}, our model has a
consistent performance when we change the distribution at test time, despite being trained
exclusively on the uniform 1-3 distribution.

\begin{table}[t]
\begin{center}
\begin{small}
\resizebox{0.9\columnwidth}{!}{%
\begin{tabular}{c|c|ccc}
\toprule
Multi-pass distribution & Oracle & LKH3  & Ours (RL)     & Ours (IL)     \\
\midrule
Uniform 1 - 3           & 1.28   & 1.80  & \textbf{1.32} & 1.42          \\
Uniform 1 - 5           & 1.92   & 3.05  & \textbf{2.11} & 2.20          \\
Uniform 1 - 10          & 3.50   & 5.72  & \textbf{3.82} & 4.24          \\
TruncGaussian 1 - 3     & 1.51   & 2.52  & \textbf{1.75} & 1.79          \\
Either 2 or 4           & 1.72   & 2.49  & \textbf{1.84} & 2.01          \\
Only 3                  & 1.52   & 1.92  & \textbf{1.68} & \textbf{1.68} \\
Exp (mean = 2)          & 1.67   & 3.26  & \textbf{1.73} & 1.84          \\
\bottomrule
\end{tabular}
}
\end{small}
\vspace{-0.1in}
\caption{Model performance on different multi-pass distributions}
\vspace{-0.2in}
\label{tab:multipass}
\end{center}
\end{table}

\vspace{-0.1in}
\paragraph{Number of value iterations:}
In this experiment, we extend the number of iterations in our value iteration module to see if the
module can benefit from longer reasoning. We originally trained our model with 5 iteration, and find
that when we scale the number of iterations up to 10 during evaluation, the performance also further
increases, as is seen in Fig.~\ref{fig:all}C.

\begin{table}[t]
\begin{center}
\begin{small}
      \resizebox{0.9\columnwidth}{!}{%
\begin{tabular}{c|cccc}
\toprule
Method                              & Network Size    & Cost            & Gap             & Runtime         \\
\midrule
Concorde (Oracle)                   & -               & 4.22            & 0.00\%          & 40.1            \\
LKH3                                & -               & 4.22            & 0.00\%          & 159             \\
OR Tools                            & -               & 4.27            & 1.11\%          & 15.0            \\
\midrule
Random Insertion                    & -               & 4.44            & 5.12\%          & \textbf{2.31}   \\
Nearest Insertion                   & -               & 4.84            & 14.7\%          & 15.4            \\
Farthest Insertion                  & -               & 4.34            & 2.36\%          & 4.66            \\
Nearest Neighbour                   & -               & 5.02            & 19.0\%          & 14.2            \\
\midrule
AM (SS)                             & 28 MB           & 4.24            & 0.51\%          & 16.3            \\
AM (SS + SP)                        & 28 MB           & 4.23            & 0.13\%          & 552             \\
MARVIN                              & 0.04 MB           & 4.54            & 7.56\%          & 61.7            \\
MARVIN (SS)                         & 0.04 MB           & 4.32            & 2.38\%          & 18.9            \\
MARVIN (SS + SP)                    & 0.04 MB           & \textbf{4.23}   & \textbf{0.10\%} & 1714            \\
\bottomrule
\end{tabular}
}
\end{small}
\caption{Single agent TSP on synthetic graphs of size 25.
We abbreviate methods that make use of \emph{self-starting} with \textbf{SS}
and \emph{sampling} with \textbf{SP}.}
\label{tab:toy}
\end{center}
\vspace{-0.2in}
\end{table}

\vspace{-0.1in}
\paragraph{Toy TSP:} To validate that our model can also solve toy TSP problems and thoroughly be
compared with previous methods under their settings, we run a single-agent TSP benchmark with graphs
of size 25 and uniformly generated vertices in a 1 by 1 square, following \citep{am}. Random,
Nearest, and Farthest insertion, as well as nearest neighbour and AM are all taken from~\cite{am}.
Usually, in our problem setting the agent has no control over where it begins. However, since a
complete TSP tour is independent of its starting position, we also test the effect of letting our
model choose its starting position, which we denote as \emph{self-starting}. We also evaluate how
other conventional tour augmentation techniques affect our model's performance.
\emph{Sampling} takes random samples from the action space of each of the agents and chooses the
one with the lowest overall cost. When augmenting the method with trajectory sampling, we sample
from 1280 model-guided stochastic runs. As shown in Table~\ref{tab:toy}, we find that even though
our model performs worse than the state-of-the-art attention model when we simply take a single greedy trajectory, it is able to outperform it
when both models are augmented with trajectory sampling, getting within 0.10\% of the optimal. It is
also worth noting that our model is 800 times smaller than the best performer in terms of the number
of parameters.

\vspace{-0.1in}
\paragraph{Visualization of large scale mapping:} We also visualize our model navigating a swam of
agents on a large portion of Chicago. Here we perform a large scale autonomous mapping simulation
with a fleet of 20 vehicles on a graph of size 2426 nodes.
We generally observe that when compared to other deep learning
solutions, our model results in a much more thorough sweep, where it more rarely has to revisit
previously seen regions. We further observe that models trained with imitation learning adopt more ``exploratory'' strategies, where
agents split from the main swarm to visit new regions. For more details on the implications of this strategy
please refer to the Appendix

\begin{table}[t]
    \begin{center}
    \begin{small}
    \resizebox{0.85\columnwidth}{!}{%
    \begin{tabular}{c|ccc|c}
    \toprule
    Variant                &  Attn. & Dense adj. & LSTM         & Action Acc.    \\
    \midrule
    GVIN                   &             &            &             & 65.4\%          \\
    GAT                    & \checkmark  &            &             & 23.5\%          \\
    No LSTM                & \checkmark  & \checkmark &             & 71.7\%          \\
    Full                   & \checkmark  & \checkmark & \checkmark  & \textbf{75.8\%} \\
    \bottomrule
    \end{tabular}
    }
    \end{small}
    \caption{Action prediction accuracy on different value iteration module designs.
    All models are trained using imitation learning.}
    \vspace{-0.25in}
    \label{tab:accuracy}
    \end{center}
\end{table}
\vspace{-0.1in}
\paragraph{Value iteration module:}
We investigated various design choices of the value iteration module, shown in
Table~\ref{tab:accuracy}, where we train different modules using IL and test them in
terms of action prediction accuracy. As shown, GVIN, lacking the dense adjacency matrix and
attention mechanism, is significantly worse than our model, and adding an LSTM further improves the
performance by allowing an extended number of iterations of value reasoning.

\vspace{-0.1in}
\paragraph{Communication module:}
We investigated different potential designs of the communication module,
including CommNet style, MaxPooling, and the number of channels. As shown in Fig.~\ref{fig:all}D, we found that our attention based
modules performs significantly better. Furthermore, the performance is improved with more channels.

\vspace{-0.1in}
\section{Conclusion}

In this paper we proposed a novel approach to perform online routing of a swarm of agents in the
realistic domain where dynamic challenges are present. By making use of learned value iteration
transitions and an attention based communication protocol, our model is able to outperform the
state-of-the-art on real road graphs. Furthermore, it is able to do so in a scalable manner and can generalize well to different number of agents and nodes without re-training.
Future work include performing a more in-depth analysis on the
information encoded in the communication and its semantic meaning.
There is also future exploration to be done into techniques to enable this system to run on massive
graphs, as the
memory usage during evaluation is still $O(n^2)$ due to the pairwise adjacency matrix.

\section*{Acknowledgement} We thank Siva Manivasagam and Abbas Sadat for helpful discussions.

\bibliography{marvin}
\bibliographystyle{icml2020}

\cleardoublepage
\appendix
\section{Model Action Visualization}

We firstly perform a qualitative analysis as to what kind of decisions each agent makes, and what
these decisions look like at the value function level. We then we visualize the paths that a fleet of
agents takes when mapping a large portion of Chicago from a bird's-eye-view perspective. We use this to
again qualitatively compare the strategies exploited by MARVIN trained with RL, MARVIN trained with IL,
and the GVIN.

\subsection{Value Function Heat Map}

\begin{figure*}[t]
  \begin{center}
    \begin{tabular}{llll}
      \includegraphics[height=3.0cm,trim={0.2cm 0 0.4cm 0},clip]{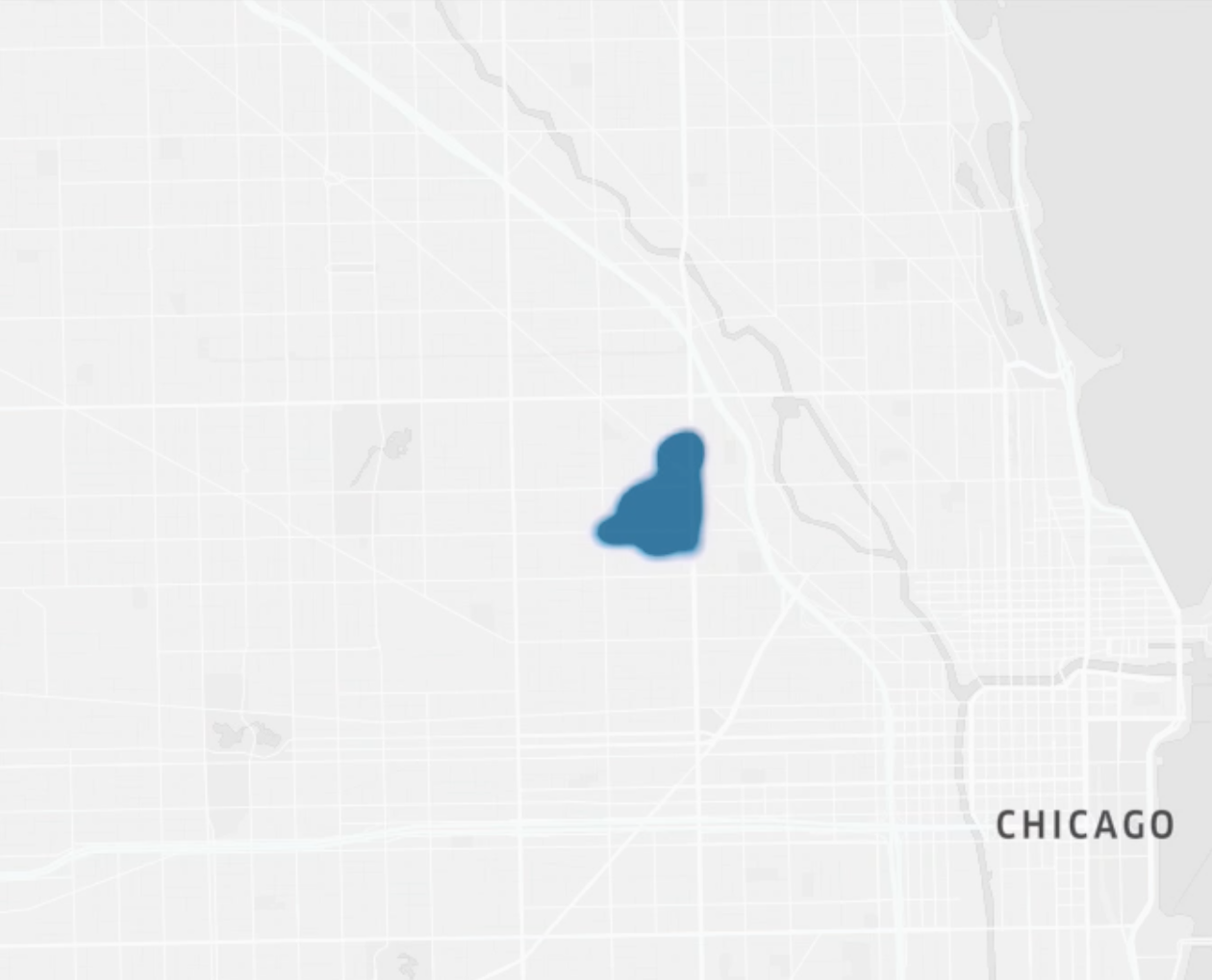} &
      \includegraphics[height=3.0cm,trim={0.2cm 0 0.8cm 0},clip]{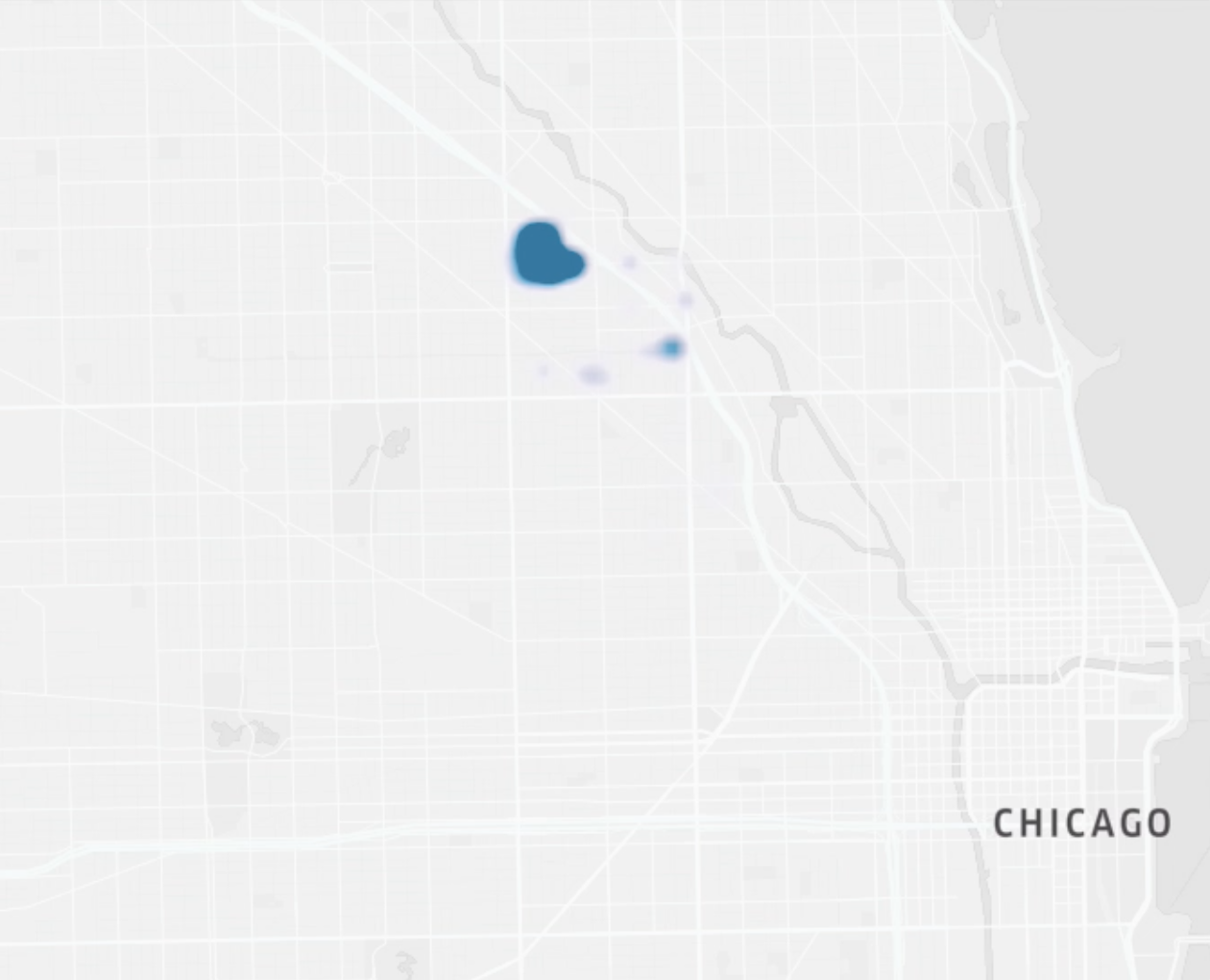} &
      \includegraphics[height=3.0cm,trim={0.2cm 0 0.8cm 0},clip]{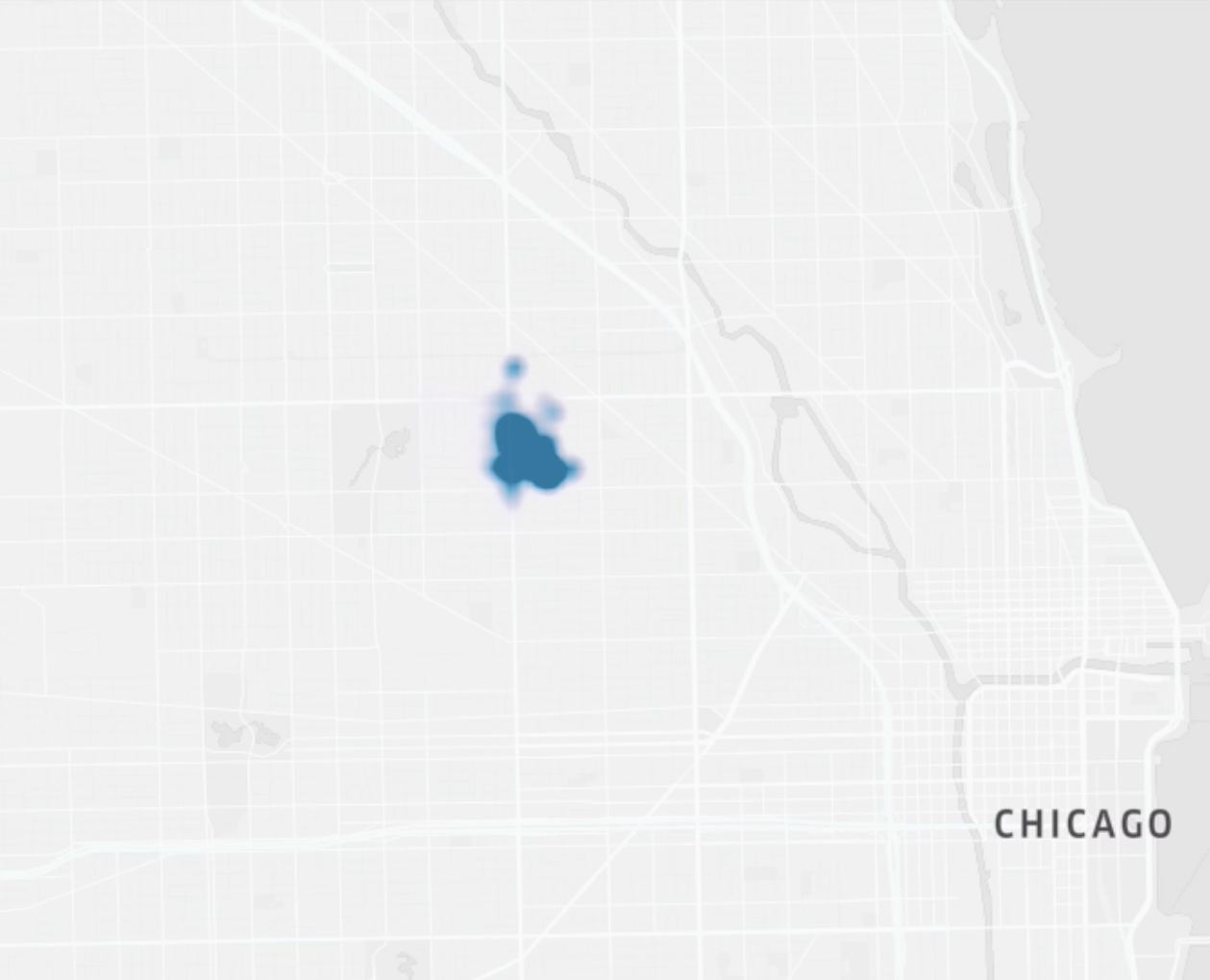} &
      \includegraphics[height=3.0cm,trim={0.0cm 0 0cm 0},clip]{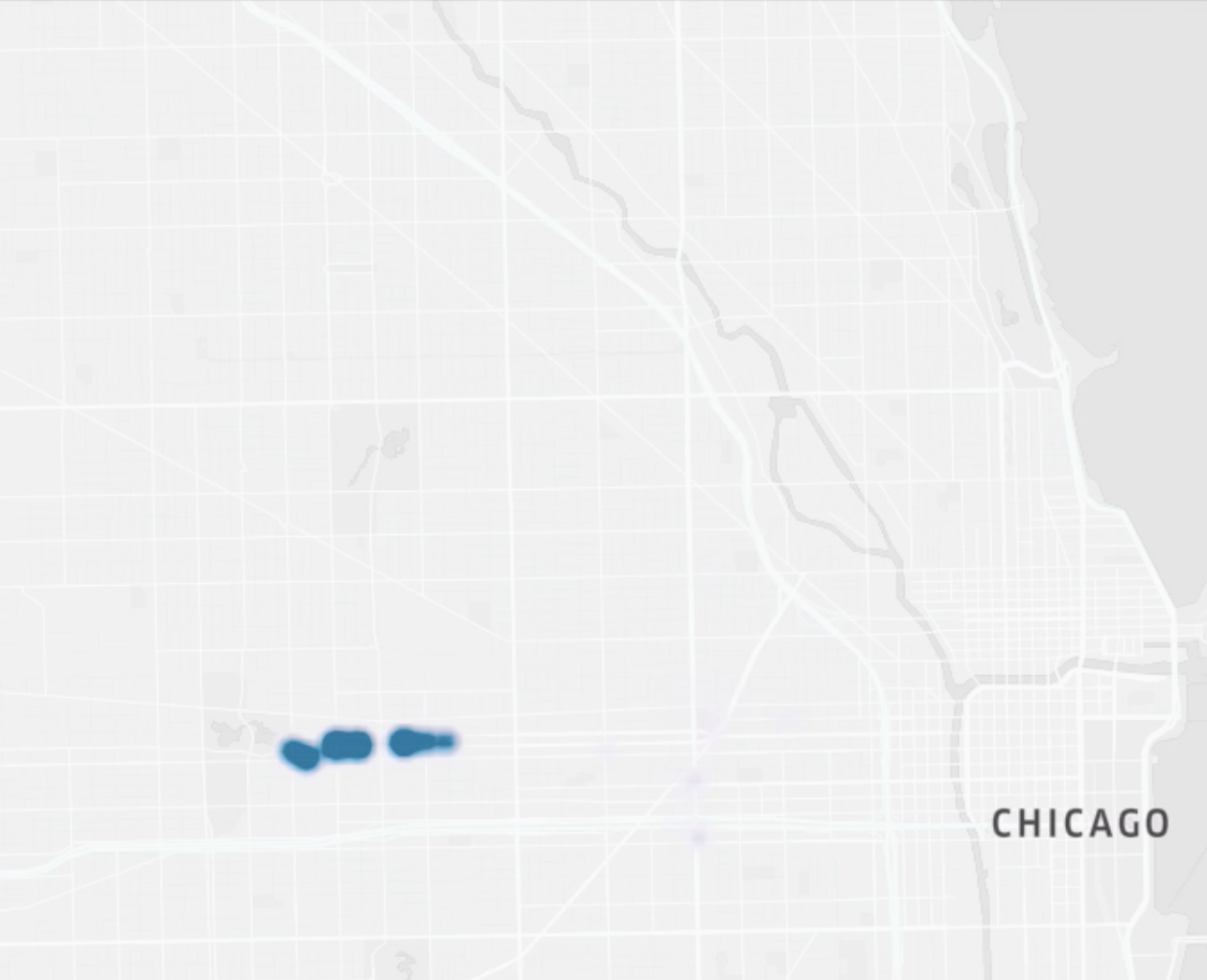} \\
    \end{tabular}
  \caption{Value function of an agents while mapping a given region.
  The value function is high around the roads that are close
  to the agent in a sense that implies that this cluster has been
  assigned to that agent. The red dot on the left and the blue dot
  on the right represent the agent's current position.}
  \label{fig:heatmap_cluster}
  \end{center}
  \vspace{-0.25in}
\end{figure*}

Upon performing an analysis of the value function, we find that two main behaviors are observed. The
first is that the agent localizes the highest points in its value function to a small region of
unvisited streets. This is equivalent to having this small cluster assigned to the agent
by the collective swarm, and then it sequentially visiting all the streets until they receive a new
objective or finish with this cluster, as can be see in Figure~\ref{fig:heatmap_cluster}.

\begin{figure*}[t]
  \begin{center}
    \begin{tabular}{llll}
      \includegraphics[height=3.0cm,trim={0.2cm 0 0.4cm 0},clip]{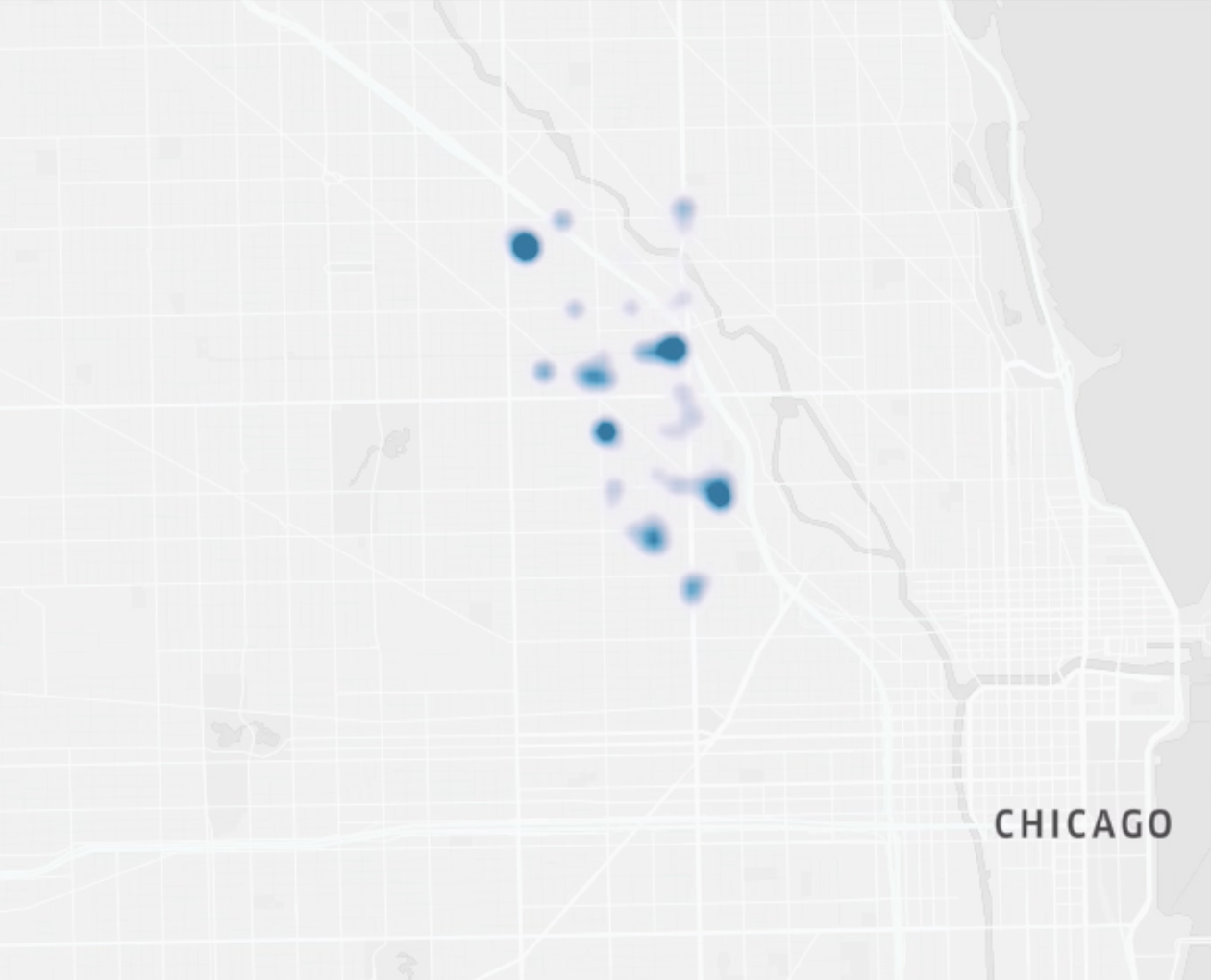} &
      \includegraphics[height=3.0cm,trim={0.2cm 0 0.8cm 0},clip]{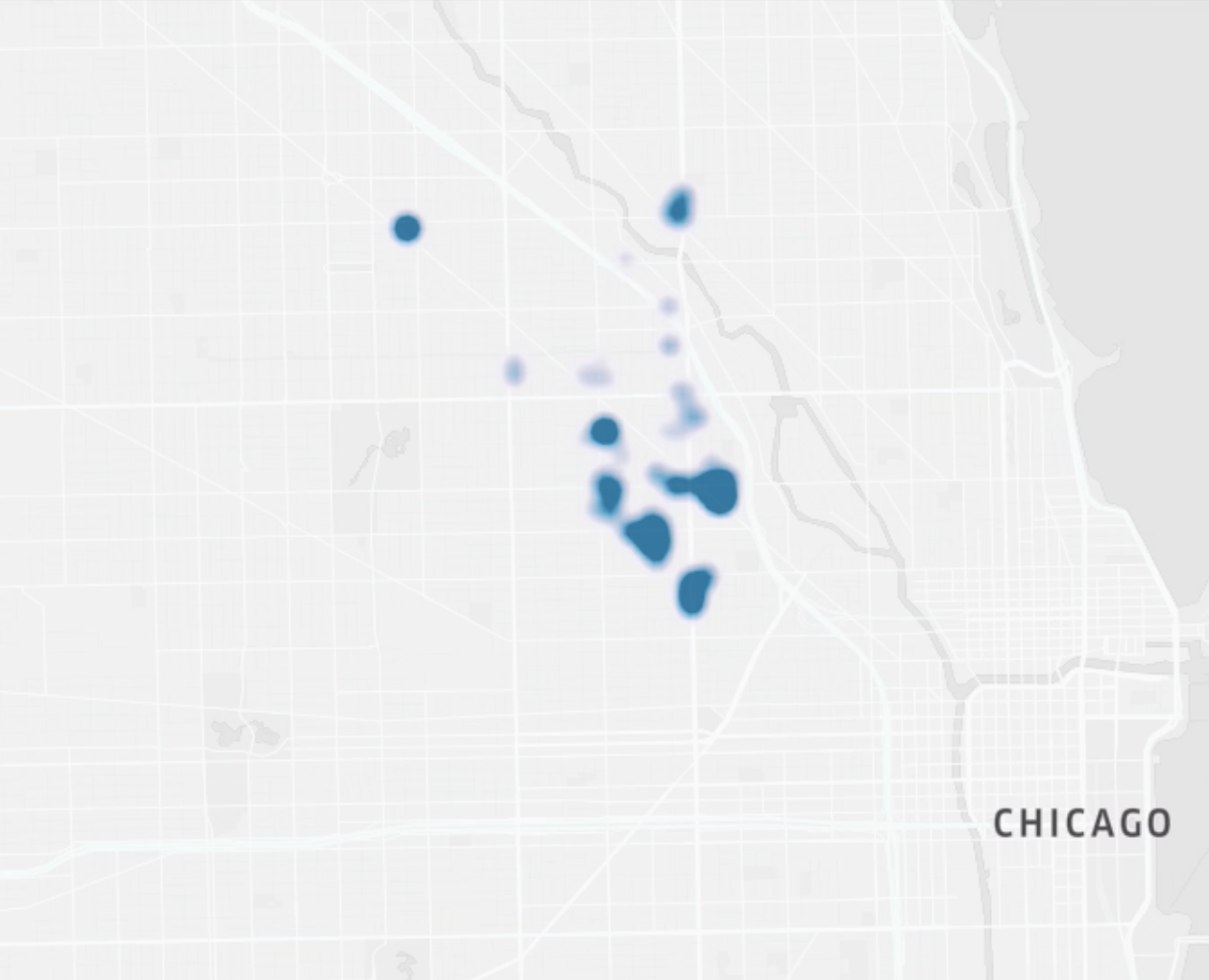} &
      \includegraphics[height=3.0cm,trim={0.2cm 0 0.8cm 0},clip]{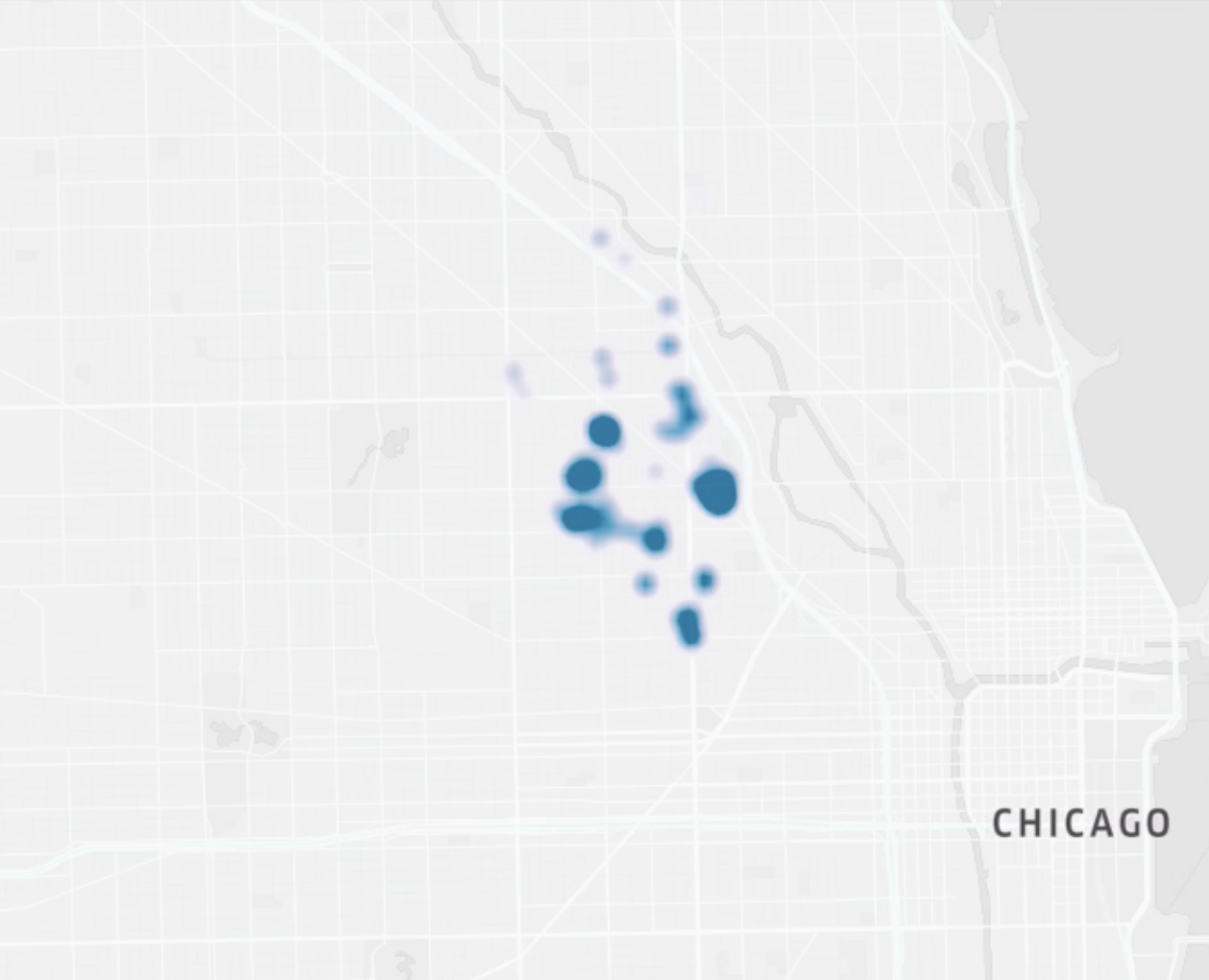} &
      \includegraphics[height=3.0cm,trim={0.0cm 0 0cm 0},clip]{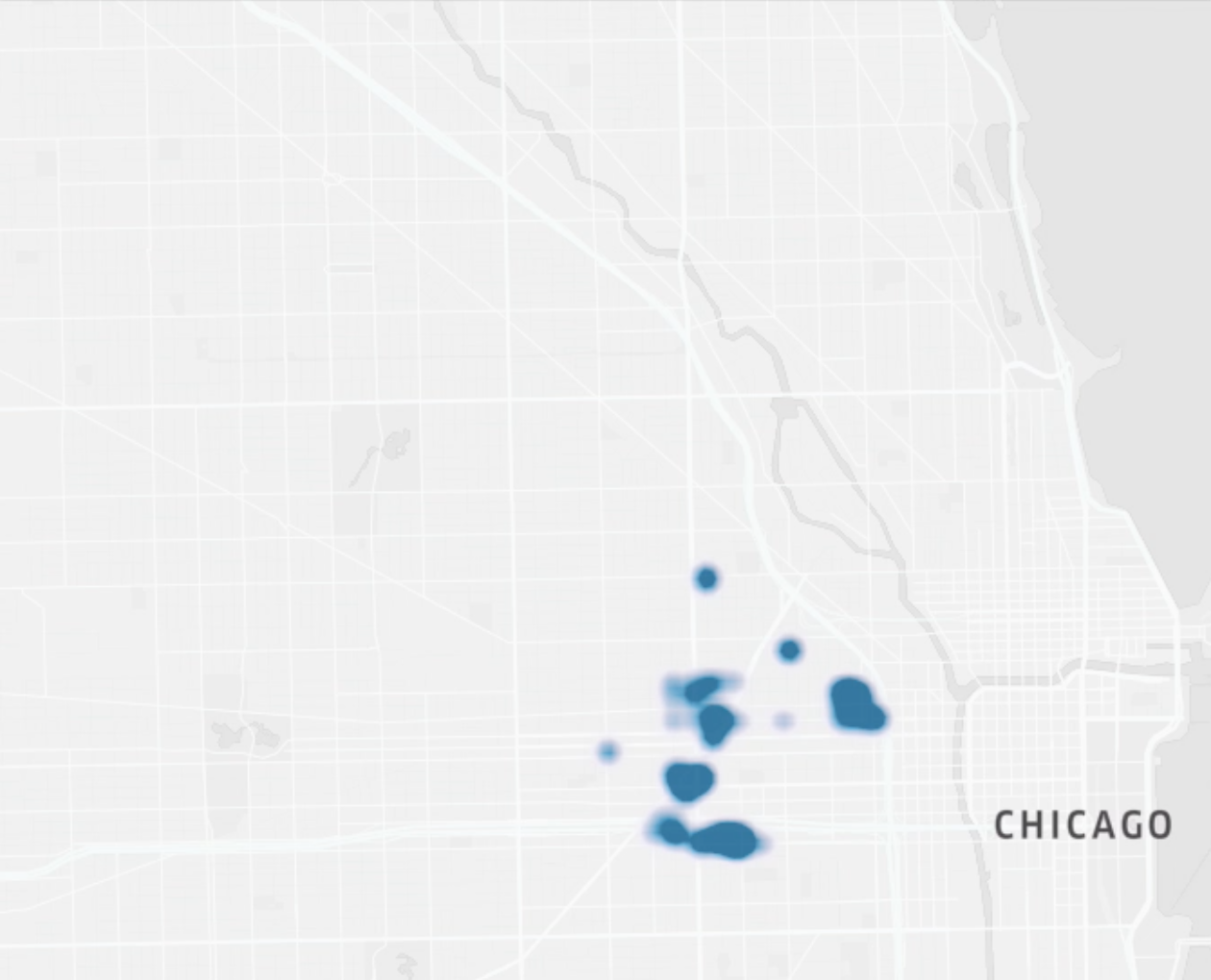} \\
    \end{tabular}
  \caption{Value function of an agents while mapping a given region.
  The value function is high around the roads that are distant and sporadically
  distributed implying an exploratory strategy. The red dot on the left and the blue dot
  on the right represent the agent's current position.}
  \label{fig:heatmap_sporadic}
  \end{center}
  \vspace{-0.25in}
\end{figure*}

The secondary observed behavior, as seen in Figure~\ref{fig:heatmap_sporadic}, is that the agents
occasionally begin increasing their value function at far away nodes. This can be interpreted
as the exploration phase, where the agents are encouraged to travel longer
distances in order to reach new sub-clusters that need to be
mapped. The peaks of the value function also appear to be relatively sporadic, indicating the
ability of the agents to consider a wide variety of potential routes.

\subsection{Overall Swarm Strategy}

We qualitatively compare our model's overall strategy to that of the generalized value iteration
network (GVIN) and observe that in general, the GVIN network tends to promote exploration.
We also observe that this high level strategy fails to cover all streets in a reliable
manner. There are small sections throughout the graph that remain unvisited, and
in order to perform a full traversal the agents must eventually
return to these small unvisited sections, often covering significant distances in
the process. This stands in contrast to what we observe with MARVIN, where the network prioritizes
covering each street in a region before moving on to the next area. This ultimately results in less
of a need for revisited regions that have incomplete mapping.

\begin{figure*}[t]
  \begin{center}
    \begin{tabular}{ll}
      \includegraphics[height=8.1cm,trim={0.2cm 0 0.4cm 0},clip]{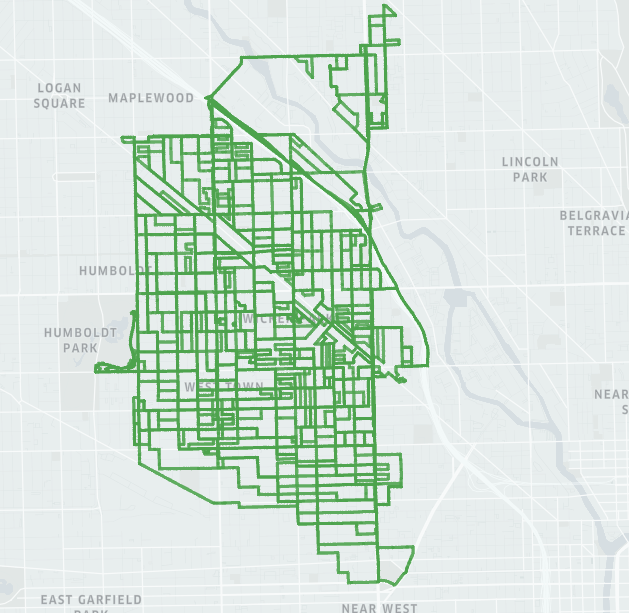} &
      \includegraphics[height=8.1cm,trim={0.2cm 0 0.8cm 0},clip]{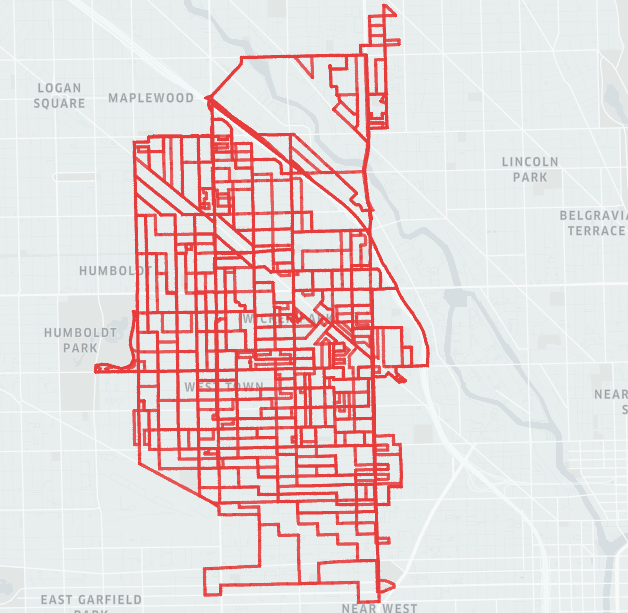} \\
    \end{tabular}
  \caption{Bird's eye view of a partially complete traversal
  of MARVIN (left) and of the GVIN (right). While slightly more
  spread out, the traversal of the GVIN leaves many small streets
  unvisited and is less thorough overall.}
  \label{fig:vin_gvin}
  \end{center}
\end{figure*}

Next, we compare the high level strategy of MARVIN when trained with reinforcement learning to that
when trained with imitation learning. We observe that in this context, both methods prioritize a
thorough traversal, but that the agents trained using imitation learning are more efficient and
therefore are able to expand to new regions much quicker than the agents trained with reinforcement
learning. This matches the trend we noted when comparing training procedures and the scalability of
the models that they produce.

\begin{figure*}[t]
  \begin{center}
    \begin{tabular}{ll}
      \includegraphics[height=9.1cm,trim={0.2cm 0 0.4cm 0},clip]{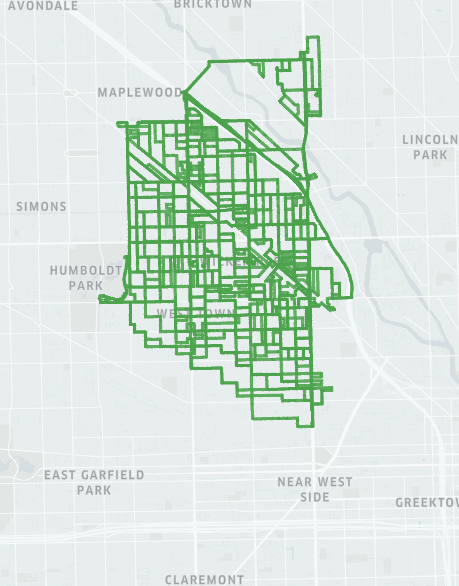} &
      \includegraphics[height=9.1cm,trim={0.2cm 0 0.8cm 0},clip]{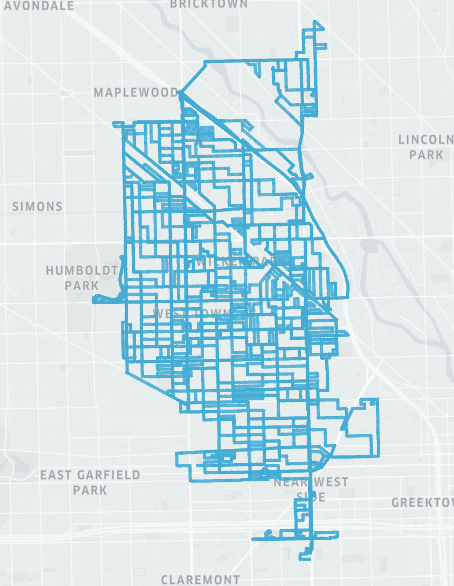} \\
    \end{tabular}
  \caption{Bird's eye view of a partially complete traversal
  of MARVIN trained with RL (left) and MARVIN trained with IL
  (right). Both are relatively thorough while expanding to new
  regions, but the model trained using imitation learning is
  able to cover the regions in a more efficient manner.}
  \label{fig:il_rl}
  \end{center}
\end{figure*}

\section{Sample Graph Visualization}

\begin{figure*}[t]
  \begin{center}
  \begin{tabular}{lll}
    \includegraphics[height=5.1cm,trim={0.2cm 0 0.4cm 0},clip]{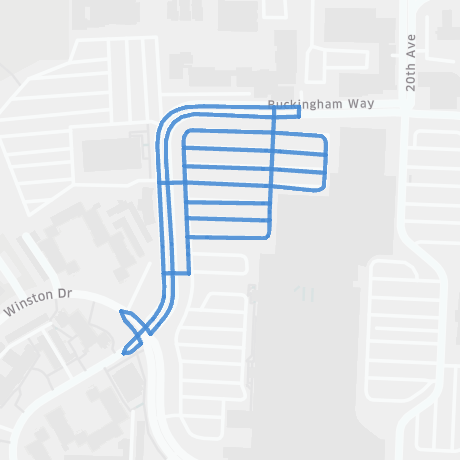} &
    \includegraphics[height=5.1cm,trim={0.2cm 0 0.4cm 0},clip]{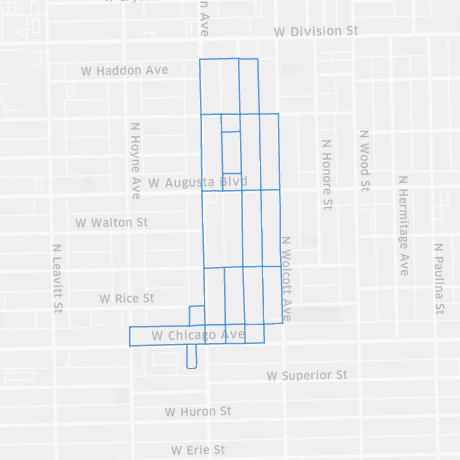} &
    \includegraphics[height=5.1cm,trim={0.2cm 0 0.4cm 0},clip]{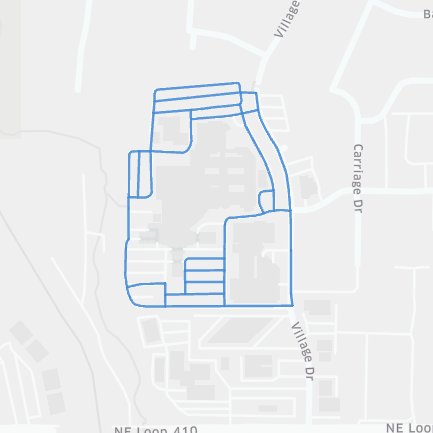} \\
    \includegraphics[height=5.1cm,trim={0.2cm 0 0.4cm 0},clip]{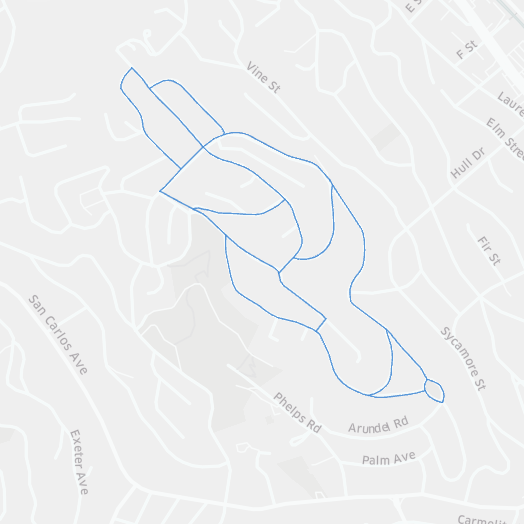} &
    \includegraphics[height=5.1cm,trim={0.2cm 0 0.4cm 0},clip]{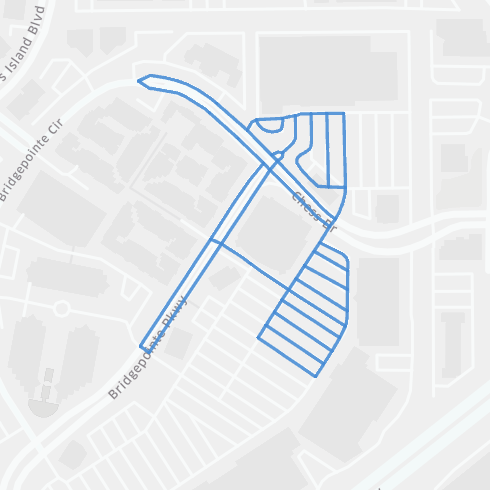} &
    \includegraphics[height=5.1cm,trim={0.2cm 0 0.4cm 0},clip]{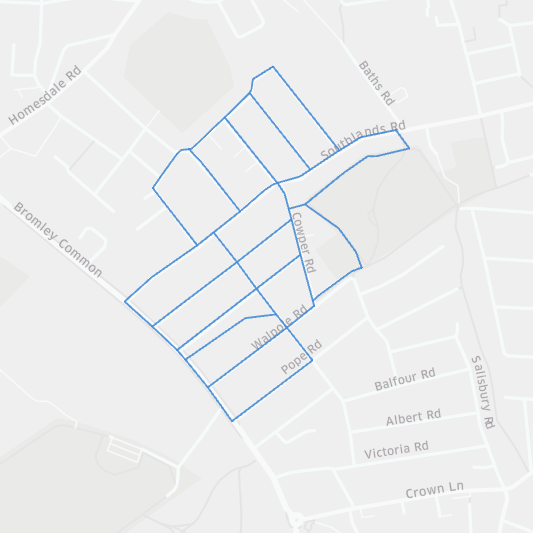} \\
  \end{tabular}
  \caption{Random graphs sampled from the training set. }
  \label{fig:example_graphs}
  \end{center}
  \vspace{-0.25in}
\end{figure*}

We visualize a few of the graphs that are used in the training process seen in Figure \ref{fig:example_graphs}.
While each one can be represented by a strongly connected graph, they nevertheless possess distinct
features which enable a higher degree of generalization during the training process.

\end{document}